\documentclass{article}
\usepackage{ijcai26}

\usepackage{times}
\usepackage{soul}
\usepackage{url}
\usepackage[utf8]{inputenc}
\usepackage[small]{caption}
\usepackage{graphicx}
\usepackage{amsmath}
\usepackage{amsthm}
\usepackage{amssymb}
\usepackage{booktabs}
\usepackage{algorithm}
\usepackage{algorithmic}
\usepackage[switch]{lineno}
\usepackage{microtype}
\usepackage{multirow}
\usepackage[hidelinks]{hyperref}

\title{Absorbing Gradient Conflicts: Modeling Semantic Variance via Kent Distributions for Cross-Modal Hashing}

\author{
    Submission Number: 2392
}

\author{
    Hengjie Zhu$^{1,2}$\and
    Dayan Wu$^{1}$\thanks{Corresponding Author}\and 
    Zihao Zhang$^{1,2}$\and
    Xinze Liu$^{1,2}$\and
    Jingxuan Yu$^3$\and
    Peng Fu$^{1}$\and
    Zheng Lin$^1$\and
    Weiping Wang$^1$
    \affiliations{
    $^1$Institute of Information Engineering, Chinese Academy of Sciences\\
    $^2$School of Cyber Security, University of Chinese Academy of Sciences\\
    $^3$School of Cyber Science and Engineering, Southeast University\\
}
    \emails{
    \{zhuhengjie, wudayan, zhangzihao, liuxinze, fupeng, linzheng, wangweiping\}@iie.ac.cn,
    yujingxuan24@seu.edu.cn
}
}
\begin{document}

\maketitle

\begin{abstract}
Supervised proxy-based deep cross-modal hashing has become the dominant paradigm for large-scale retrieval. However, prevalent methods model class proxies as deterministic points in the embedding space. This rigid assumption causes severe gradient conflicts in multi-label scenarios, 
where gradient conflicts arising from label co-occurrence lead to severe gradient contention and optimization collapse.
To resolve this, we propose Kent-based Distributional Proxy Hashing (KDPH), a novel framework that shifts proxy representation from static points to flexible anisotropic Kent distributions on the hypersphere. Unlike point proxies that must shift their positions to accommodate conflicting gradients, KDPH absorbs these conflicts by dynamically adjusting its directional variance. This allows the proxy to maintain a stable semantic mean direction while stretching to cover diverse label correlations. Furthermore, to ensure stable training of these geometric parameters, we derive a tailored loss function incorporating the Cayley transform to enforce strict orthogonality. To the best of our knowledge, KDPH is the first framework to successfully introduce the Kent distributions into cross-modal hashing.
Experiments on three benchmark datasets demonstrate that KDPH mitigates proxy collapse and chaotic oscillation, significantly outperforms state-of-the-art methods. Code is available at https://github.com/Senmo996/KDPH-official-code.

\end{abstract}

\section{Introduction}
\begin{figure}[t!]
\centering
\includegraphics[width=0.95\columnwidth]{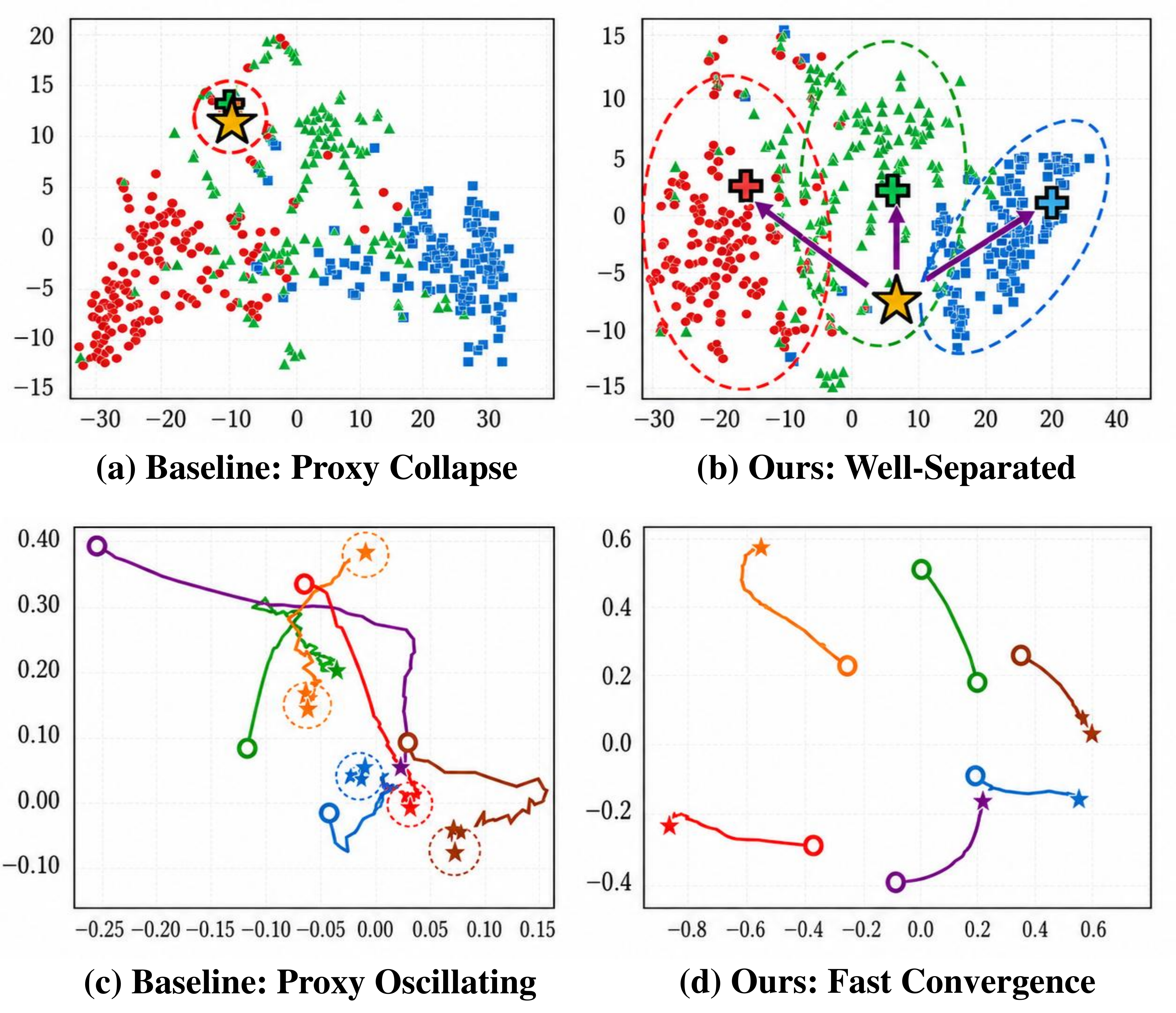}
\caption{\textbf{Comparison of proxy optimization behaviors.} 
\textbf{(a)-(b)} Conventional proxy learning suffers from \textit{proxy collapse}, where learned proxies move toward the anchor and become poorly separated, whereas KDPH preserves distinct proxy distributions. 
\textbf{(c)-(d)} Conventional methods exhibit slow and oscillatory optimization trajectories, while KDPH achieves smoother trajectories and faster convergence. Colors denote different categories.}
\label{fig:intro}
\end{figure}

Driven by the surge in multimodal data \cite{xu2024reliable}, Cross-Modal Hashing Retrieval has emerged as a standard method for large-scale information retrieval \cite{cao2022image,singh2022learning,yan2025cross}. By projecting high-dimensional features into a compact Hamming space, this approach circumvents the prohibitive computational and storage costs of conventional Euclidean methods, enabling efficient similarity approximation via bitwise XOR operations \cite{wang2022high,lu2019online,sun2022deep}. Despite rapid progress towards Supervised Deep Cross-Modal Hashing Retrieval—leveraging deep neural networks and semantic supervision for end-to-end representation learning—fundamental challenges still persist. Real-world data are inherently multi-faceted, with single instances often associated with multiple overlapping labels. This multi-label characteristic poses severe difficulties for existing supervised frameworks, which struggle to capture complex, non-exclusive semantic correlations, necessitating more robust modeling strategies.

% While significant progress has been made, fundamental limitations persist in multi-label scenarios, especially within methods relying on proxy-based strategies \cite{DSPH}. Prevalent methods typically abstract each semantic category to a singular, rigid proxy point within the embedding space. However, while effective for single-label tasks, this rigid point-based assumption induces critical optimization issues when confronted with multi-label data. The primary bottleneck is gradient contention, which stems from the complex semantic structure where a sample belongs to multiple classes simultaneously. In such a point-proxy framework, a single sample concurrently exerts pulling forces on divergent proxies, yet these proxies—modeled as rigid points with zero spatial variance—lack the capacity to accommodate such conflicting semantic signals. Consequently, this unresolved contention drives the model towards two degenerate states: (1) Chaotic Proxy Oscillation, where the proxy lacking the volumetric capacity to represent diverse contextual variations, is forced to constantly shift its position to satisfy conflicting gradients, thereby preventing the model from anchoring to a stable semantic center. (2) Proxy Collapse, where the optimization process forcibly pulls proxies of distinct categories together to minimize the aggregate loss. Specifically, proxies of low-frequency tail classes often collapse onto those of high-frequency head classes. This topological collapse eliminates fine-grained decision boundaries, severely impairing the model's discriminative power.

Despite the dominance of proxy-based strategies in deep cross-modal hashing \cite{DCPH,DSPH}, fundamental limitations persist in multi-label scenarios. The prevailing paradigm typically abstracts each semantic category as a singular, deterministic point \cite{teh2020proxynca++,Kim_2020_CVPR,Movshovitz-Attias_2017_ICCV,Qian_2019_ICCV} within the embedding space. While effective for single-label tasks, this rigid zero-variance assumption provokes severe gradient contention when confronted with the complex semantic conflicts inherent in label co-occurrence. Specifically, in a point-proxy framework, a multi-label sample simultaneously exerts pulling forces on divergent proxies. Modeled as rigid points, these proxies lack the spatial capacity to accommodate such conflicting semantic signals. Consequently, this unresolved contention drives the model towards optimization collapse. These two limitations manifesting in two degenerate states: (1) \textit{Chaotic Proxy Oscillation}, where the proxy, unable to satisfy conflicting gradients, is forced to constantly shift its position, preventing the model from anchoring to a stable semantic center; and (2) \textit{Proxy Collapse}, where the optimization process forcibly pulls proxies of distinct categories together to minimize aggregate loss—particularly causing low-frequency tail classes to collapse onto high-frequency head classes—thereby eliminating fine-grained decision boundaries and impairing discriminative power.

To address these challenges, we propose the Kent-based Distributed Proxy Hashing (KDPH) framework. We depart from the single-point paradigm by modeling each class proxy as a learnable Kent distribution—an anisotropic spherical probability model defined by a mean direction and an elliptical covariance structure. Unlike rigid point proxies that must drastically shift position to satisfy conflicting gradients, this distributed representation resolves conflicts by changing specific directional variances, thereby endowing each class with a semantic shape and volume to flexibly absorb the contextual variance introduced by label co-occurrence. 
Specifically, observing the anisotropic geometry of the Kent distribution, we design a tailored loss function to achieve gradient decoupling mechanism by stretching shape rather than moving centroid. This approach effectively resolves conflict, preventing chaotic oscillation and proxy collapse.
Furthermore, by incorporating the Cayley transform to enforce strict orthogonality for the principal axes and utilizing this probabilistic module solely to guide the learning process, KDPH achieves state-of-the-art performance with zero additional computational overhead during inference, successfully resolving the persistent trade-off between retrieval accuracy and efficiency.

\begin{figure*}[t]
\centering
\includegraphics[width=\textwidth]{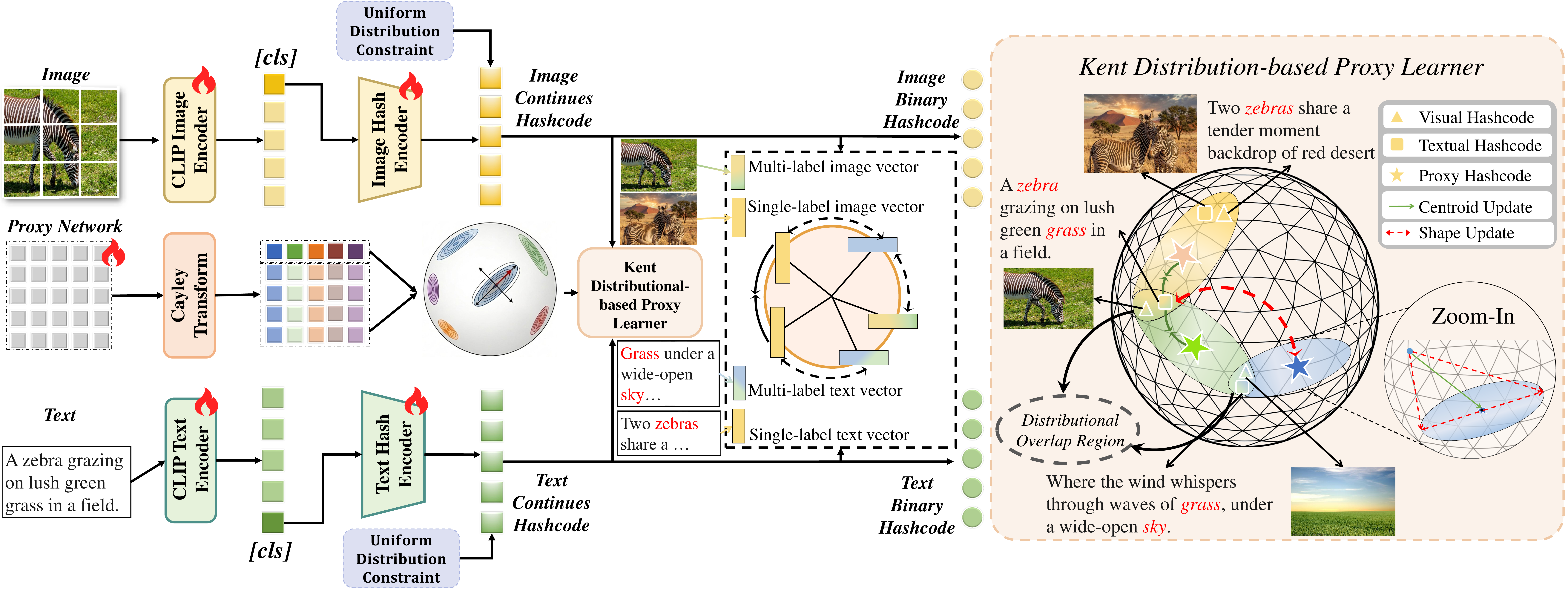}
\caption{\textbf{Overview of the Kent-based Distributed Proxy Hashing (KDPH) framework.} It consists of a CLIP-based feature extraction module and a Kent-based  Distributional Proxy Learner. As shown in the right panel, we model class proxies as learnable anisotropic distributions rather than static points. Through the gradient decoupling mechanism by separating centroid $\mu$ and variance $B, \Gamma$, the model dynamically stretches the proxy's semantic volume to accommodate multi-label variance, effectively mitigating proxy collapse and oscillation.}
\label{fig:overview}
\end{figure*}

In summary, our contributions are as follows:
\begin{itemize}
    % \item We are the first to shift traditional point proxy to anisotropic Kent distributions in cross-modal hashing, fundamentally resolve severe gradient contention and optimization collapse in multi-label scenarios.
    \item We introduce KDPH, a novel framework that reformulates proxy-based hashing by substituting rigid point proxies with anisotropic Kent distributions. To the best of our knowledge, KDPH is the first work to demonstrate that modeling proxy uncertainty through directional distributions can eliminate the optimization collapse caused by gradient contention in multi-label cross-modal retrieval.
    % \item We derive a tailored loss function incorporating Cayley transform, which allows proxies to dynamically absorb gradient conflicts, successfully eliminating \textbf{Proxy Collapse} and \textbf{Chaotic Proxy Oscillation}.
    \item We derive a manifold-constrained loss function leveraging the Cayley transform, which enables proxies to adaptively absorb gradient conflicts via directional variance rather than centroid displacement, thereby effectively eradicating \textbf{Proxy Collapse} and suppressing \textbf{Chaotic Proxy Oscillation} during the learning process.
    \item Extensive experiments demonstrate that KDPH achieves \textbf{state-of-the-art} performance on three benchmarks, empirically validating the efficacy of our gradient decoupling strategy and showing superior robustness particularly in complex multi-label scenarios.
\end{itemize}

\section{Related Work}

\subsection{Supervised Deep Cross-Modal Hashing and Proxy Learning}
Supervised deep cross-modal hashing has advanced significantly, positioned such as CNN-RNN and adversarial baselines \cite{DCMH,SSAH,DADH}. Recently, graph and transformer-based models \cite{GCH,DCHMT,mith,ijcai2024p69,liu2024dual} have further pushed the boundaries of performance. 
In parallel, these models improving their algorithmic design from simple pairwise constraints \cite{DNpH} to adaptive gradient-triplet losses \cite{DAGtH} or sophisticated geometric \cite{DScPH} strategies which ranging from spherical mutual information to boundary-sensitive mining \cite{DDBH}—to refine the Hamming space structure. 
To address the scalability bottlenecks of pairwise comparisons, proxy-based metric learning has been effectively incorporated. By representing categories as distinct point proxy, this paradigm allows the model to efficiently organize the embedding space \cite{DCPH,DSPH}, and has recently explored hyperbolic geometries for hierarchical modeling \cite{DHaPH}. However, a major challenge persists that the deterministic point formulation's inherent zero-variance assumption fundamentally limits its applicability to multi-label scenarios with high contextual variance.

\subsection{Probabilistic Embeddings and Non-Isotropy}
Probabilistic embedding learning has emerged as a robust alternative to deterministic point estimates~\cite{Lin_2018_ECCV}. To respect the unique geometry of hyperspherical feature spaces, the field has evolved from early Euclidean Gaussian models~\cite{shi2019probabilistic} to spherical probability density functions, such as the von Mises-Fisher (vMF) distribution~\cite{Zhe2018,Park2019,Li2021}. However, the efficacy of vMF-based approaches remains constrained by their inherent isotropy.  One major challenge is the implicit assumption of uniform variance in all directions and this rigid simplification fails to accommodate the complex, anisotropic semantic manifolds inherent in multi-label data. Although recent metric learning approaches have explored non-isotropy through lightweight mechanisms, such as variance scaling or geometric regularization~\cite{roth2022non,kirchhof2022non}, these low-parameter heuristics often lack the sufficient geometric flexibility required for large-scale multi-label cross-modal hashing, where semantic manifolds are densely entangled. To bridge this gap, we propose the KDPH, which introduces the parameter-rich Kent distribution to deliberately increase geometric capacity. We further facilitate stable training by resolving gradient contention and optimization collapse via the Cayley transform and a tailored loss function.

% Probabilistic embedding learning has emerged as a pivotal direction for capturing data uncertainty, positioned as a robust alternative to deterministic point estimates~\cite{Lin_2018_ECCV}. To respect the unique geometry of hyperspherical feature spaces, the field has evolved from early Euclidean Gaussian models~\cite{shi2019probabilistic} to spherical probability density functions, such as the von Mises-Fisher (vMF) distribution~\cite{Zhe2018,Park2019,Li2021}. However, the efficacy of vMF-based approaches remains constrained by their inherent isotropy. One major challenge is the implicit assumption of uniform variance in all directions; this rigid simplification fails to accommodate the complex, anisotropic semantic manifolds inherent in multi-label data. While recent metric learning approaches have explored non-isotropy through lightweight mechanisms, such as variance scaling or geometric regularization~\cite{roth2022non,kirchhof2022non}, these low-parameter heuristics often lack the sufficient geometric flexibility required for large-scale multi-label cross-modal hashing, where semantic manifolds are densely entangled. Moreover, merely constraining the loss function is insufficient to explicitly construct the necessary decision boundaries. To this end, we propose the KDPH framework, which introduces the parameter-rich Kent distribution to deliberately increase geometric capacity. We further facilitate stable training by resolving gradient contention and optimization collapse via the Cayley transform and a tailored loss function.

\section{Method}
\subsection{Formulation and Architecture}

Given a training dataset $O = \{(\boldsymbol{i}_n, \boldsymbol{t}_n, \boldsymbol{l}_n)\}_{n=1}^N$ with images $\boldsymbol{i}_n$, texts $\boldsymbol{t}_n$, and multi-label vectors $\boldsymbol{l}_n \in \{0, 1\}^C$, we extract $D$-dimensional semantic features $\boldsymbol{f}_n^i$ and $\boldsymbol{f}_n^t$ via a dual-stream Transformer backbone initialized with pre-trained CLIP. Modality-specific hashing networks $H_I, H_T: \mathbb{R}^D \rightarrow \mathbb{R}^K$ then project these features into $K$-dimensional continuous representations $\boldsymbol{h}_n^i$ and $\boldsymbol{h}_n^t$. The final binary hash codes are obtained via the sign function:
\begin{equation}
    \resizebox{.91\linewidth}{!}{$
        \displaystyle
        \boldsymbol{b}_n^i := \text{sign}(\boldsymbol{h}_n^i) \in \{-1, 1\}^K,
        \quad
        \boldsymbol{b}_n^t := \text{sign}(\boldsymbol{h}_n^t) \in \{-1, 1\}^K
    $}
\end{equation}

\subsection{Kent-based Distributed Proxy Mechanism}

Conventional hashing methods typically abstract each class as deterministic point whose rigid zero-variance assumption fundamentally unsuitable for multi-label scenarios, and conflicting semantic signals from co-occurring labels lead to gradient contention and optimization collapse. To resolve this, KDPH shifts the paradigm from static points to distributional proxies on the hypersphere \cite{wang2020understanding}. By modeling each class as a flexible Kent distribution with learnable centroid and variance, our framework effectively absorbs the gradients variance arising from label co-occurrence.

\subsubsection{Kent Distribution for Proxy Modeling}

To instantiate this distributional paradigm, we employ the generalized Kent distribution. This strategic choice addresses the inherent limitations of simpler spherical models, such as the von Mises-Fisher (vMF) distribution. While vMF offers mathematical convenience, its isotropic nature rigidly imposes uniform variance across all dimensions, rendering it ill-suited for capturing the complex, directional manifolds typical of multi-label features. In contrast, the Kent distribution is intrinsically anisotropic. It provides the necessary geometric flexibility to model elliptical contours on the hypersphere $\mathbb{S}^{K-1}$ , where $K$ denotes the hash code dimension, distinct from the feature dimension $D$.

Formally, the probability density function (PDF) for a class $c$ is defined as:
\begin{equation}
    \resizebox{.91\linewidth}{!}{$ 
        \displaystyle
        p(\boldsymbol{z} | \boldsymbol{\Theta}_c) = \frac{1}{c(\kappa,B)} \exp\left( \kappa_c \boldsymbol{\mu}_c^T \boldsymbol{z} + \sum_{i=1}^{N_{axes}} \beta_{c,i} (\boldsymbol{\gamma}_{c,i}^T \boldsymbol{z})^2 \right)
    $}
\end{equation}

where $c(\kappa,B)$ denotes the normalizing constant. The distribution is parameterized by $\boldsymbol{\Theta}_c = \{\boldsymbol{\mu}_c, \kappa_c, \Gamma_c, B_c\}$, in which each component serves a distinct semantic role: the mean direction $\boldsymbol{\mu}_c \in \mathbb{S}^{K-1}$ acts as the semantic centroid representing the stable core identity of the class, while $\kappa_c$ governs the global concentration of samples around this centroid. Crucially, to model non-uniform variance, $\Gamma_c$ defines the orthogonal principal axes, and $B_c = \{\beta_{c,i}\}$ controls the anisotropic shaping along these dimensions. By learning distinct $\beta_{c,i}$ values, the proxy can dynamically stretch or compress along specific semantic directions. This mechanism allows the distribution to accommodate the high semantic variance introduced by multi-label co-occurrences without drastically shifting the centroid $\boldsymbol{\mu}_c$.

\subsection{Hash learning}
To jointly optimize the deep networks and proxy parameters $\boldsymbol{\Theta}_c$ while respecting manifold geometry, we introduce a differentiable reparameterization technique. This mechanism is integrated with a Kent-based log-likelihood scoring function and a multi-task objective to strictly enforce both semantic discrimination and quantization constraints.

\subsubsection{Orthogonal Frame Construction via Cayley Transform}
Directly optimizing $\boldsymbol{\mu}_c$ and $\boldsymbol{\Gamma}_c$ is challenging due to the strict orthogonality required by the Kent distribution. To address this, we construct a unified orthonormal basis using the Cayley Transform. Specifically, for each class $c$, we learn a skew-symmetric parameter matrix $\mathbf{A}_c \in \mathbb{R}^{K \times K}$. By exploiting the mapping between the Lie algebra $\mathfrak{so}(K)$ and the Lie group $SO(K)$, we generate a strict rotation matrix $\mathbf{Q}_c$:
\begin{equation}
  \mathbf{Q}_c = (\mathbf{I} + \mathbf{A}_c)^{-1}(\mathbf{I} - \mathbf{A}_c) \in SO(K)  
\end{equation}

This transformation ensures that $\mathbf{Q}_c$ remains orthogonal regardless of the updates to $\mathbf{A}_c$. We then define the proxy components by slicing $\mathbf{Q}_c$: the first column serves as the centroid $\boldsymbol{\mu}_c$, while the following columns form the shape axes $\boldsymbol{\Gamma}_c$.  This parameterization naturally enforces geometric constraints, guaranteeing orthogonality not only between the centroid and the axes but also pairwise among all principal axes within $\boldsymbol{\Gamma}_c$, effectively bypassing the need for expensive orthogonalization procedures.

\subsubsection{Anisotropic Energy Scoring and Parameterization}
To evaluate the semantic affinity between a query sample and the class proxies, we first project the continuous hash code $\boldsymbol{h}$ onto the unit hypersphere, $\boldsymbol{z} = \boldsymbol{h} / \|\boldsymbol{h}\|_2 \in \mathbb{S}^{K-1}$, ensuring geometric consistency with the directional statistics.

While the canonical Kent distribution defines a rigorous probability density function, its optimization is hindered by a computationally intractable normalization constant dependent on high-dimensional Bessel functions. Since our primary objective in hashing is discriminative ranking rather than exact density estimation, we circumvent this bottleneck by reinterpreting the log-density formulation as a geometric Compatibility Score. We define the anisotropic energy score $S(\boldsymbol{z}, c)$ for a class $c$ as follows:
\begin{equation}
S(\boldsymbol{z}, c) = \kappa_c (\boldsymbol{\mu}_c^T \boldsymbol{z}) + \sum_{i=1}^{N_{\text{axes}}} \beta_{c,i} (\boldsymbol{\gamma}_{c,i}^T \boldsymbol{z})^2 - \mathcal{R}(\kappa, \boldsymbol{\beta})
\end{equation}
Here, we employ a logarithmic regularization term $\mathcal{R}(\kappa, \boldsymbol{\beta}) = \log(\kappa_c + \|\boldsymbol{\beta}_c\|_1)$. This term serves as a soft magnitude constraint, effectively transforming the probabilistic objective into a flexible quadratic energy metric on the hypersphere, preserving the topological benefits of Kent geometry while avoiding the gradient instability of Bessel approximations.

To strictly enforce the definition of concentration and magnitude, the parameters $\kappa_c$ and $\beta_{c,i}$ are constrained to be positive via a Softplus activation on learnable scalars $\hat{\kappa}_c$ and $\hat{\beta}_{c,i}$. Crucially, regarding the shape constraints, while the canonical Kent distribution requires $2|\beta_{c,i}| < \kappa_c$ for uni-modality, we adopt a generalized Fisher-Bingham formulation by removing this upper bound. By allowing the network to learn $\boldsymbol{\beta}_c$ freely, the proxy gains the flexibility to model complex, potentially multi-modal distributions. This relaxation is particularly advantageous for capturing the high variance and distinct semantic manifolds inherent in multi-label data.

\subsubsection{Distribution-Aware Proxy Triplet Loss}
To strictly enforce discriminative margins within the hyperspherical manifold, we employ a structure-preserving Distribution-Aware Proxy Triplet Objective. Unlike standard cross-entropy which treats classes as independent points, this loss explicitly optimizes the anisotropic energy alignment between samples and learnable Kent distributions. By maximizing the energy score for relevant classes while minimizing it for irrelevant ones, the objective ensures that an instance resides firmly within the high-density acceptance region of its ground-truth semantic distributions, effectively pushing decision boundaries away from the ambiguous overlapping zones.

Formally, given a normalized query sample $\boldsymbol{z}$, a positive proxy distribution $c_p \in \mathcal{P}$ (where $\mathcal{P}$ is the set of ground-truth labels), and a negative proxy $c_n \in \mathcal{N}$, we enforce a strict separation constraint based on the energy score $S(\cdot)$:
\begin{equation}
   S(\boldsymbol{z}, c_p) > S(\boldsymbol{z}, c_n) + m 
\end{equation}
where $m$ is a fixed margin hyperparameter governing the minimum semantic gap.

To accommodate the multi-label setting where a sample is associated with multiple semantic centroids, the final loss aggregates these constraints over all valid triplets. This formulation allows the gradients to dynamically adjust the shape parameters $\boldsymbol{\beta}_c$ of the Kent distributions, absorbing the conflict from hard negatives:
\begin{equation}
    \resizebox{.91\linewidth}{!}{$
        \displaystyle
        \mathcal{L}_{Proxy} = \frac{1}{|\mathcal{P}| |\mathcal{N}|} \sum_{y \in \mathcal{P}} \sum_{j \in \mathcal{N}} \max\bigl(0, S(\boldsymbol{z}, c_n) - S(\boldsymbol{z}, c_p) + m\bigr)
    $}
\end{equation}

\begin{table*}[t]
\centering
\footnotesize
\renewcommand{\arraystretch}{0.9}
\resizebox{\textwidth}{!}{ 
\begin{tabular}{lllccccccccc}
\toprule
\multirow{2}{*}{Task} & \multirow{2}{*}{Method} & \multirow{2}{*}{Reference} & \multicolumn{3}{c}{MIRFLICKR-25K} & \multicolumn{3}{c}{NUS-WIDE} & \multicolumn{3}{c}{MS COCO} \\
\cmidrule(lr){4-6} \cmidrule(lr){7-9} \cmidrule(lr){10-12} 
 & & & 16bits & 32bits & 64bits & 16bits & 32bits & 64bits & 16bits & 32bits & 64bits \\
\midrule
\multirow{11}{*}{I2T} 
 & DCPH   & TKDE'22 & 0.7679 & 0.7733 & 0.7764 & 0.6161 & 0.6156 & 0.6180 & 0.5206 & 0.5806 & 0.5939 \\
 & DCMHT  & MM'22   & 0.8263 & 0.8272 & 0.8441 & 0.6799 & 0.6992 & 0.7038 & 0.6447 & 0.6757 & 0.6915 \\
 & nivMF  & ECCV'22 & 0.8241 & 0.8245 & 0.8352 & 0.7023 & 0.6902 & 0.7024 & 0.7075 & 0.7242 & 0.7523 \\
 & MIAN   & TKDE'22 & 0.8123 & 0.8220 & 0.8355 & 0.6130 & 0.5894 & 0.6057 & 0.5350 & 0.5579 & 0.5404 \\
 & DSPH   & TCSVT'23& 0.8129 & 0.8482 & 0.8541 & 0.6830 & 0.6979 & 0.7162 & 0.7044 & 0.7510 & 0.7694 \\
 & DNPH   & MC'24   & 0.8108 & 0.8269 & 0.8229 & 0.6689 & 0.6811 & 0.6939 & 0.6438 & 0.6910 & 0.7294 \\
 & DNpH   & TMM'24  & 0.8423 & 0.8552 & 0.8558 & 0.6921 & 0.7022 & 0.7071 & 0.6727 & 0.6903 & 0.6860 \\
 & RDPH   & TMM'24  & 0.7420 & 0.7749 & 0.7867 & 0.6330 & 0.6317 & 0.6523 & 0.5867 & 0.7151 & 0.7368 \\
 & DDBH   & TCSVT'24& 0.8450 & 0.8534 & 0.8610 & 0.7041 & 0.7145 & 0.7229 & 0.7165 & 0.7454 & 0.7681 \\
 & DNcH   & ESWA'25 & --     & --     & --     & 0.7150 & 0.7305 & 0.7387 & 0.7199 & 0.7341 & 0.7509 \\
 & DAGtH  & ESWA'25 & 0.8268 & 0.8514 & 0.8599 & 0.6827 & 0.6882 & 0.6927 & --     & --     & --     \\
 & KDPH   & Ours    & \textbf{0.8770} & \textbf{0.8639} & \textbf{0.8639} & \textbf{0.7430} & \textbf{0.7365} & \textbf{0.7392} & \textbf{0.7636} & \textbf{0.7761} & \textbf{0.7843} \\ 
\midrule
\multirow{11}{*}{T2I} 
 & DCPH   & TKDE'22 & 0.7992 & 0.8046 & 0.8061 & 0.6187 & 0.6292 & 0.6394 & 0.5981 & 0.5879 & 0.5745 \\
 & DCMHT  & MM'22   & 0.8116 & 0.8137 & 0.8281 & 0.6876 & 0.7104 & 0.7253 & 0.6513 & 0.6832 & 0.7052 \\
 & nivMF  & ECCV'22 & 0.8125 & 0.8102 & 0.8213 & 0.6912 & 0.6952 & 0.6924 & 0.6421 & 0.6892 & 0.7042 \\
 & MIAN   & TKDE'22 & 0.8058 & 0.8151 & 0.8182 & 0.6365 & 0.6312 & 0.6416 & 0.5489 & 0.5493 & 0.5365 \\
 & DSPH   & TCSVT'23& 0.8000 & 0.8238 & 0.8294 & 0.6997 & 0.7153 & 0.7304 & 0.7040 & 0.7556 & 0.7713 \\
 & DNPH   & MC'24   & 0.8015 & 0.8176 & 0.8166 & 0.6871 & 0.6994 & 0.7182 & 0.6468 & 0.7012 & 0.7388 \\
 & DNpH   & TMM'24  & 0.8147 & 0.8292 & 0.8361 & 0.6992 & 0.7137 & 0.7139 & 0.6562 & 0.6860 & 0.6928 \\
 & RDPH   & TMM'24  & 0.7200 & 0.7504 & 0.7713 & 0.6732 & 0.7067 & 0.7190 & 0.5925 & 0.6910 & 0.7139 \\
 & DDBH   & TCSVT'24& 0.8245 & 0.8318 & 0.8390 & 0.7194 & 0.7211 & 0.7325 & 0.7167 & 0.7394 & 0.7595 \\
 & DNcH   & ESWA'25 & --     & --     & --     & 0.7196 & 0.7343 & 0.7430 & 0.7130 & 0.7306 & 0.7398 \\
 & DAGtH  & ESWA'25 & 0.8072 & 0.8270 & 0.8382 & 0.6936 & 0.7042 & 0.7098 & --     & --     & --     \\
 & KDPH   & Ours    & \textbf{0.8351} & \textbf{0.8439} & \textbf{0.8480} & \textbf{0.7473} & \textbf{0.7399} & \textbf{0.7458} & \textbf{0.7627} & \textbf{0.7785} & \textbf{0.7795} \\
\bottomrule
\end{tabular}
}
\caption{mAP comparisons on three benchmark datasets with references. The best results are highlighted in bold.}
\label{tab:comparison_sota_full}
\end{table*}

% This objective drives the optimization of the Kent proxies to maximize inter-class separability while maintaining intra-class compactness.

\subsubsection{Multi-Modal Irrelevance Regularization}

To strictly separate semantically disjoint samples, we incorporate the Multi-Modal Irrelevance Loss \cite{DSPH}. This objective penalizes high similarities between irrelevant pairs—defined as instances sharing no common labels (i.e., $l_i \cdot l_j = 0$). Taking the image intra-modal regularization as an example, the loss is defined as the average cosine similarity over all such pairs in a batch:
\begin{equation}
\mathcal{L}_{reg\_i} = \frac{\sum_{i=1}^{B} \sum_{j=1}^{B} \mathbb{I}(l_i \cdot l_j = 0) \cos(\boldsymbol{h}_i^x, \boldsymbol{h}_j^x)}{\sum_{i=1}^{B} \sum_{j=1}^{B} \mathbb{I}(l_i \cdot l_j = 0)}
\end{equation}
where $\mathbb{I}(\cdot)$ is the indicator function and $B$ is the batch size. The text intra-modal loss $\mathcal{L}_{reg\_t}$ and the cross-modal loss $\mathcal{L}_{reg\_it}$ are computed analogously by substituting the continues hash feature pairs with $(\boldsymbol{h}^y, \boldsymbol{h}^y)$ and $(\boldsymbol{h}^x, \boldsymbol{h}^y)$, respectively. The final regularization term aggregates these three components: $\mathcal{L}_{Reg} = \mathcal{L}_{reg\_i} + \mathcal{L}_{reg\_t} + \mathcal{L}_{reg\_it}$.

\subsubsection{Cross-Modal Contrastive Alignment}

To bridge the heterogeneity between modalities, we employ the symmetric InfoNCE \cite{he2020momentum} objective to project representations into a shared, modality-invariant hypersphere. This contrastive loss maximizes the mutual information between matched image-text pairs. The image-to-text alignment loss is formulated as:
\begin{equation}
\mathcal{L}_{i \to t} = -\frac{1}{B} \sum_{k=1}^{B} \log \frac{\exp(\cos(\boldsymbol{h}_k^x, \boldsymbol{h}_k^y) / \tau)}{\sum_{j=1}^{B} \exp(\cos(\boldsymbol{h}_k^x, \boldsymbol{h}_j^y) / \tau)}
\end{equation}
where $\tau$ is a learnable temperature parameter scaling the distribution sharpness. The text-to-image loss $\mathcal{L}_{t \to i}$ is computed symmetrically by swapping the anchor and positive samples. The final cross-modal alignment objective is the sum of both directions: $\mathcal{L}_{CM} = \mathcal{L}_{i \to t} + \mathcal{L}_{t \to i}$.

\subsubsection{Uniform Distribution Constraint for Hashing}

To maximize code entropy and prevent bit-collapse, we enforce a Uniform Distribution Constraint via Optimal Transport. Specifically, we minimize the Wasserstein-1 distance between the learned batch codes $\boldsymbol{B}^x$ and a sampled ideal uniform distribution $\boldsymbol{A} \in \{-1, 1\}^{B \times K}$. This is formulated as finding the optimal permutation matrix $\boldsymbol{O}^x$ that minimizes the transport cost:
\begin{equation}
\mathcal{L}_{U,x} = \min_{\boldsymbol{O}^x \in \mathcal{O}} -\text{tr}(\boldsymbol{O}^x \boldsymbol{A} (\boldsymbol{B}^x)^T)
\end{equation}
where $\mathcal{O}$ is the set of all $B \times B$ permutation matrices. The constraint is applied symmetrically to both modalities, yielding the total loss $\mathcal{L}_U = \mathcal{L}_{U,x} + \mathcal{L}_{U,y}$.

\subsubsection{Overall Objective Function}
Our comprehensive training objective $\mathcal{L}_{Total}$ is a multi-task function, designed to integrate the four distinct components discussed previously. It simultaneously optimizes the model for (1) proxy learning, (2) cross-modal alignment, (3) semantic irrelevance separation, and (4) high-entropy hash code generation.The final objective is a weighted summation of these components:
\begin{equation}
\mathcal{L}_{Total} = \mathcal{L}_{Proxy} + \epsilon \mathcal{L}_{Reg} + \zeta \mathcal{L}_{CM} + \eta \mathcal{L}_{U}
\end{equation}

where $\epsilon$, $\zeta$, and $\eta$ are scalar hyperparameters that balance the 
contribution of each term.

\section{Experiment}
To demonstrate the validity of our proposed KDPH framework, we conducted comprehensive experiments on three publicly available cross-modal multi-label datasets MIRFLICKR-25K, NUS-WIDE and MS COCO. Besides, we introduce the datasets for the experiments and explain the details of the implementation of KDPH and evaluate metrics.
\begin{figure}[t]
\centering
\begin{minipage}{\linewidth}
\centering
% 将 \linewidth 改为 0.8\linewidth 来缩小图片
\includegraphics[width=\linewidth]{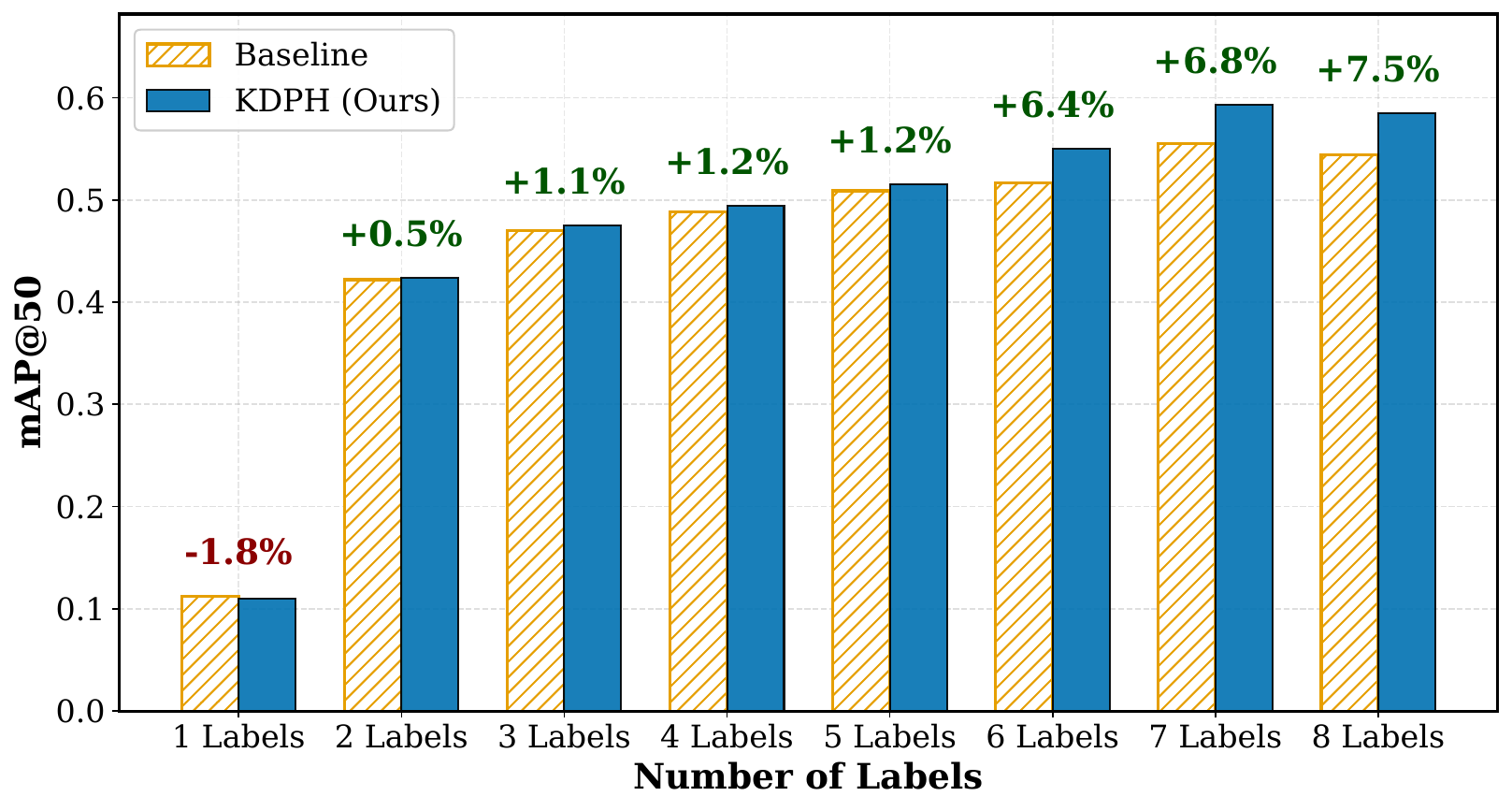}
\vspace{-0.5em}
\caption*{(a) Retrieval Performance vs. Semantic Complexity}
\end{minipage}

\vspace{0.1cm} % 增加上下两张图之间的垂直间距

\begin{minipage}{\linewidth}
\centering
% 将 \linewidth 改为 0.8\linewidth 来缩小图片
\includegraphics[width=\linewidth]{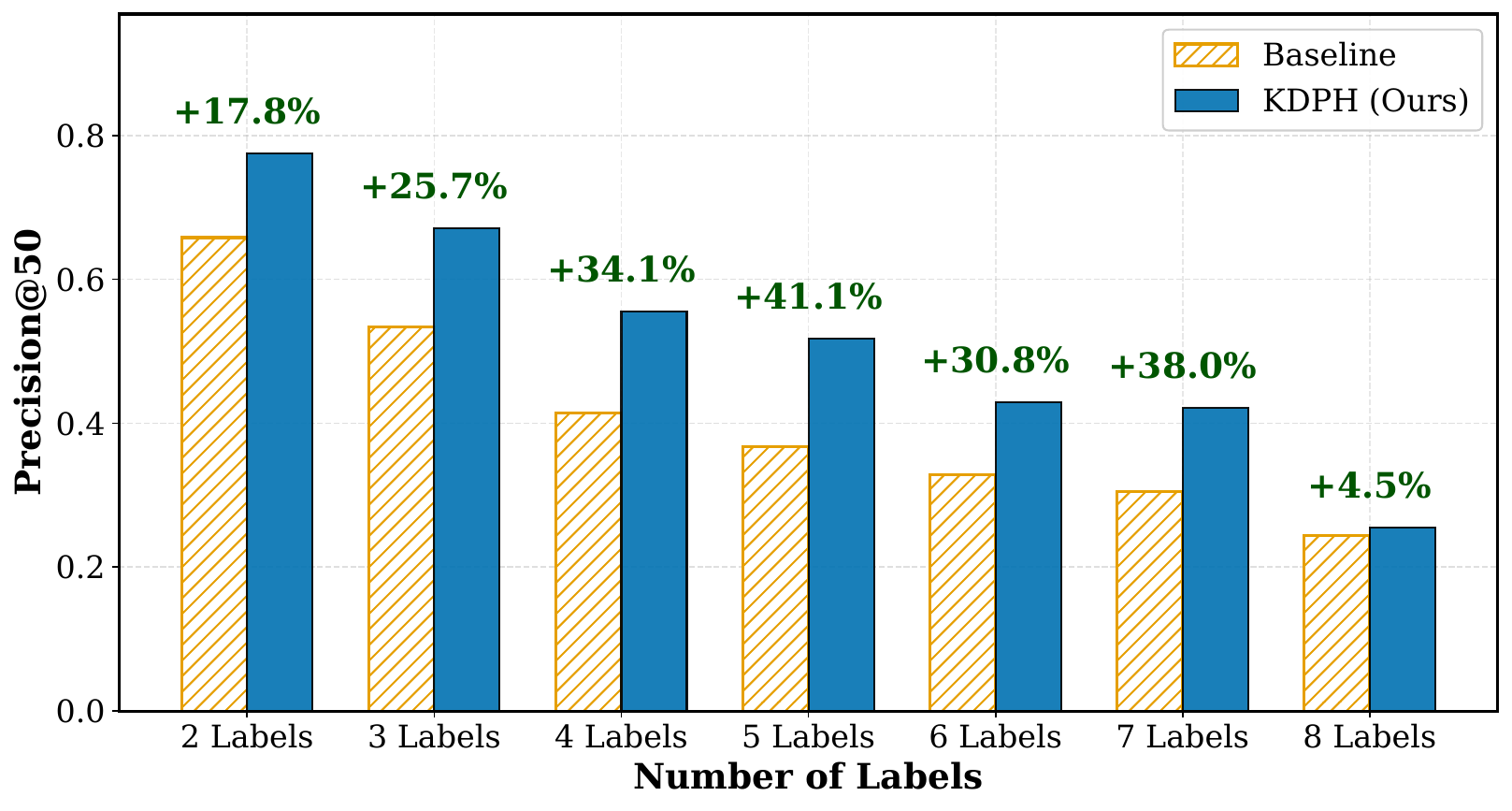}
\vspace{-0.5em}
\caption*{(b) Precision of 'Hard-to-Find' Labels}
\end{minipage}
\caption{Impact of Semantic Complexity on Model Performance.}
\label{fig:collapse_analysis}
\end{figure}

\subsection{Experimental Settings}
\subsubsection{Datasets}
We evaluate KDPH on three standard multi-label datasets. MIRFLICKR-25K \cite{mirflickr} contains 24,581 image-text pairs across 24 classes. NUS-WIDE \cite{chua2009nus} comprises 269,648 pairs; following standard protocols, we utilize a subset of 195,834 pairs from 21 frequent categories. MS COCO \cite{lin2014coco} consists of 123,289 images, each with 5 captions, spanning 80 categories. Across all datasets, we randomly sample 5,000 pairs as the query set and use the remainder as the database, from which 10,000 pairs are uniformly sampled to construct the training set.

\begin{figure}[tb]
\centering
\includegraphics[width=\linewidth]{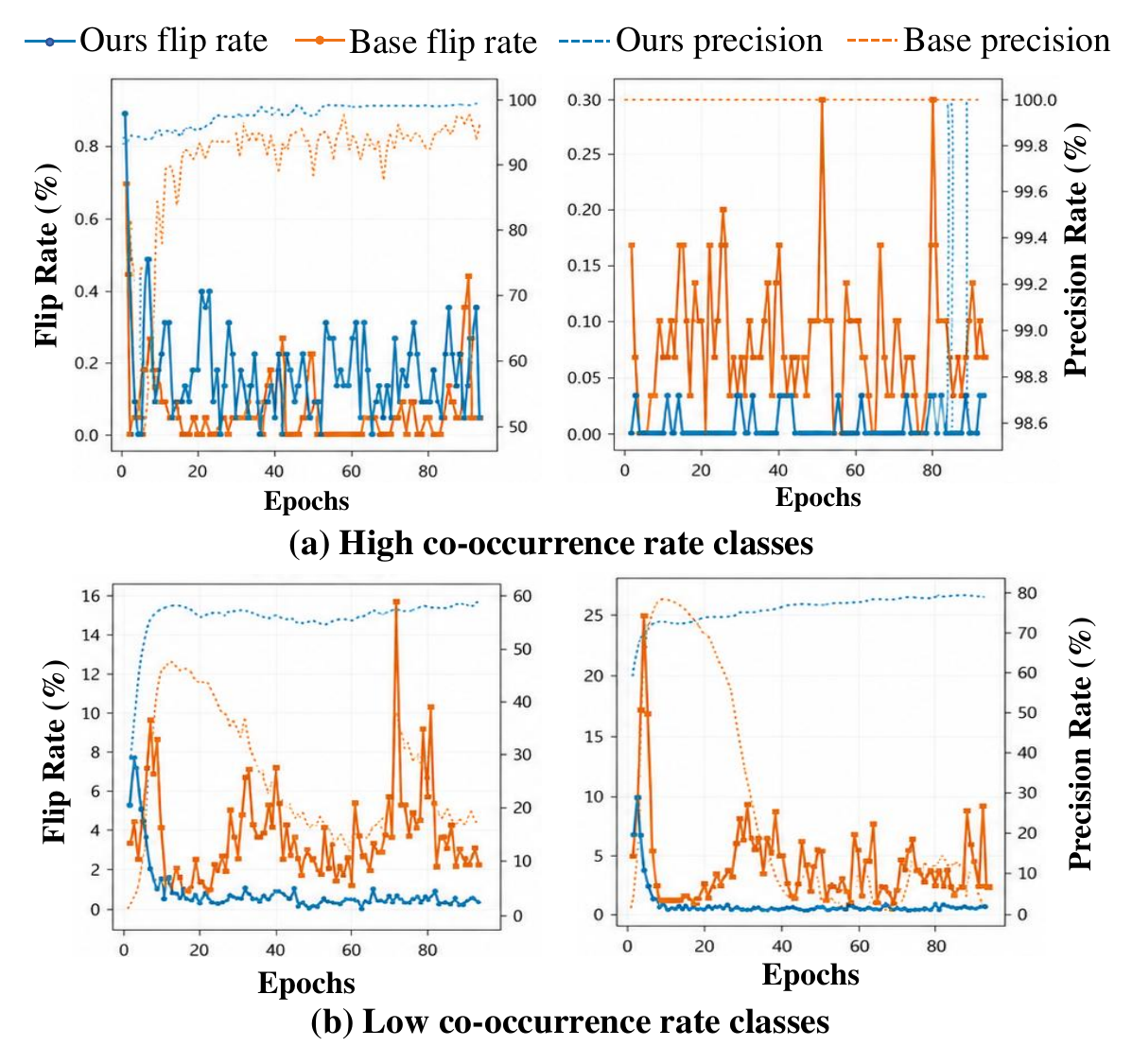}
\caption{Training dynamics comparing DSPH and KDPH.}
\label{fig:oscillation_analysis}
\end{figure}

\subsubsection{Baseline Methods}
In our experiments, we selected 9 state-of-the-art deep cross-modal hashing methods for comparison, which contain DCPH\cite{DCPH}, nivMF\cite{kirchhof2022non}, DCHMT\cite{DCHMT}, MIAN\cite{MIAN}, DSPH\cite{DSPH}, MITH\cite{mith}, DNPH\cite{DNPH_SS}, DNpH\cite{DNpH}, DDBH\cite{DDBH}, DNcH\cite{DNcH} and DAGtH\cite{DAGtH}.

\subsubsection{Experimental Details}
We implemented KDPH in PyTorch on an NVIDIA A800 GPU, optimizing via Adam (LR=0.001, batch size=128). The hyper-parameters are set as $\epsilon=40$, $\zeta=0.3$, $\eta=0.1$. Regarding the anisotropic geometry, we set the number of principal axes as $N_{axes}=K/8$, where $K$ denotes the target hash code length (e.g., $N_{axes}=8$ for 64-bit codes). To ensure a fair comparison, we follow all baselines that use the identical CLIP backbone and data splits.

\subsection{Main Results}
To evaluate the effectiveness of our KDPH framework, we conduct a comprehensive comparison with state-of-the-art baselines on three benchmark datasets. The comparative results are summarized in Table \ref{tab:comparison_sota_full}. The proposed KDPH method outperforms all state-of-the-art baselines with significant performance improvements across all hash code lengths. This phenomenon demonstrates that the anisotropic Kent distributions possess stronger feature extraction and discriminative capabilities than the deterministic point-based baselines.In particular, our KDPH yields superiority over nivMF, a representative approach employing non-isotropic von Mises-Fisher distributions. It is observed that such methods suffer from the limitations of geometric inflexibility, particularly in scenarios enriched with complex semantic entanglements. This limitation is empirically evidenced by the substantial margin of over \textbf{7.92\%} on the challenging MS COCO dataset ($I2T$ @ 16 bits).Despite the reliable performance achieved by previous methods, our KDPH benefits from the Kent-based distributional proxy and modeling elliptical semantic variance, demonstrates superior performance.

Furthermore, to verify the performance of our KDPH under highly compressed spaces, we investigate the Bit-Efficiency. In comparison to the state-of-the-art methods (e.g., DCPH, DSPH) which suffer from sharp performance degradation under limited encoding capacity, KDPH maintains high discriminative power at 16 bits. Specifically, our KDPH yields an improvement of approximately $6.07\%$ on MS COCO over the second-best method. These findings suggest that the meticulously designed Kent distributions offer superior information density, effectively absorbing gradient contentions even within highly compressed Hamming spaces. KDPH also achieves good balance between modalities, indicating that the optimization strategy successfully harmonizes visual and textual distributions within the shared latent space, thereby preventing the modality domination often observed in existing baselines.

\subsection{Analysis}

\subsubsection{Mitigation of Chaos Proxy Oscillation via KDPH}

We first investigate the robustness of our KDPH approach in preventing optimization collapse. So we monitor the track of Proxy Flip Rate $R_{flip}^{(t)}$ \cite{xie2016unsupervised}, defined as the ratio of samples changing their nearest proxy assignment between consecutive epochs. As illustrated in Figure \ref{fig:oscillation_analysis}, it can be observed that in Low Co-occurrence Classes (Fig. \ref{fig:oscillation_analysis}b), both the baseline and KDPH exhibit stable convergence, confirming that rigid point proxies suffice for simple, unimodal manifolds.In contrast, regarding High Co-occurrence Classes (Fig. \ref{fig:oscillation_analysis}a), the baseline suffers from severe oscillation and jagged precision curves, leading to a deficiency in optimization stability. Benefiting from the advantages of anisotropic Kent distributions, KDPH maintains exceptional stability with a flip rate $<1\%$. Specifically, our framework is designed to dynamically absorb gradient variance through shape deformation rather than through straightforward centroid shifting. In this way, the proposed method effectively mitigates the chaotic oscillations that plague traditional point-based methods, ensuring better adaptability to complex semantic environments.

% To evaluate optimization stability, we monitor the \textit{Proxy Flip Rate} $R_{flip}^{(t)}$ \cite{xie2016unsupervised}, defined as the ratio of samples changing their nearest proxy assignment between consecutive epochs.

% As shown in Figure \ref{fig:oscillation_analysis}, a clear dichotomy emerges based on semantic complexity. In \textbf{Low Co-occurrence Classes} (Fig. \ref{fig:oscillation_analysis}b), both the baseline and KDPH exhibit stable convergence, confirming that rigid point proxies suffice for simple, unimodal manifolds. 

% However, in \textbf{High Co-occurrence Classes} (Fig. \ref{fig:oscillation_analysis}a), the baseline suffers from severe oscillation and jagged precision curves, indicative of unresolved gradient contention. In sharp contrast, KDPH maintains exceptional stability with a flip rate $<1\%$. By leveraging anisotropic Kent distributions to \textit{absorb} contextual variance through shape deformation rather than centroid displacement, our framework effectively eliminates the chaotic oscillations that plague traditional point-based methods.

\subsubsection{Prevention of Proxy Collapse via the KDPH}

To comprehensively evaluate the effectiveness in prevention of proxy collapse, we track the \textit{Subordinate Label} as the category with the least frequency in an sample. As illustrated in Figure \ref{fig:collapse_analysis}, Baseline suffers from the limitations in high-complexity situations where rigid point proxies degrade rapidly under the pressure of dominant classes. In contrast, KDPH method consistently demonstrates the superiority and robustness over state-of-the-art competitors, yielding a significant improvement of 41.1\% in scenarios with 5 labels. This promising performance confirms that the meticulously designed anisotropic distributions effectively mitigate the collapse issue by reserving sufficient \textit{semantic variance} for tail concepts.

Despite the reliable performance achieved by the Baseline in single-label regimes due to the advantage of point proxies, it is noted that this advantage vanishes in multi-label scenarios. This slight redundancy introduced by KDPH serves as a necessary trade-off, providing the parametric flexibility required to bridge the semantic gaps between conflicting manifolds.

\begin{table}[tbhp]
\centering
\tiny
% 1. 稍微压缩行高
\renewcommand{\arraystretch}{0.7}
% 2. 压缩列间空白 (默认 6pt)
\setlength{\tabcolsep}{3pt} 
% 3. 将宽度限制为列宽的 85% (可自行调节 0.85 这个数值)
\resizebox{0.95\columnwidth}{!}{
\begin{tabular}{cccc|cc}
\toprule
\multicolumn{4}{c|}{\textbf{Components}} & \multicolumn{2}{c}{\textbf{mAP @ 64 bits}} \\
\cmidrule(r){1-4} \cmidrule(l){5-6}
\textbf{Kent} & $\mathcal{L}_{Reg}$ & $\mathcal{L}_{CM}$ & $\mathcal{L}_{U}$ & \textbf{I $\to$ T} & \textbf{T $\to$ I} \\
\midrule
% Variant 1: Point Proxy (No Kent)
 & \checkmark & \checkmark & \checkmark & 0.7615 & 0.7542 \\
% Variant 1.5: vMF (Kent with beta=0) - NEW ROW
$\beta=0$ & \checkmark & \checkmark & \checkmark & 0.7693 & 0.7628 \\
\midrule
% Variant 2: No Regularization
\checkmark & & \checkmark & \checkmark & 0.7326 & 0.7305 \\
% Variant 3: No Cross-Modal
\checkmark & \checkmark & & \checkmark & 0.7566 & 0.7585 \\
% Variant 4: No Uniform
\checkmark & \checkmark & \checkmark & & 0.7537 & 0.7574 \\
\midrule
% Full Model
\checkmark & \checkmark & \checkmark & \checkmark & \textbf{0.7843} & \textbf{0.7795} \\
\bottomrule
\end{tabular}
}
\caption{Ablation study of different components on the MS COCO.}
\label{tab:ablation}
% \vspace{-1.5em}
\end{table}

\subsubsection{Ablation Study on Model Components}
We conduct ablation studies by systematically evaluating the mAP of each component in our framework on the MS COCO dataset. The comparative results are summarized in Table \ref{tab:ablation}.
The performance metric exhibits a distinct hierarchy. Specifically, mAP decreases from 0.7693 with the isotropic vMF distribution ($\beta=0$) to 0.7615 with rigid point proxies, suggesting only a marginal advantage for distributional proxies. Crucially, our anisotropic Kent model surpasses all state-of-the-art variants, owing to its capacity to capture \textit{anisotropic} semantic conflicts, enabling effective recalibration of semantic interactions to resolve gradient contentions.Furthermore, the Multi-Modal Irrelevance Regularization $\mathcal{L}_{Reg}$ and Cross-Modal Alignment $\mathcal{L}_{CM}$ are designed to collaboratively enhance the contextual awareness and semantic capabilities, ensuring better adaptability to nuanced semantic variations while Uniform Constraint $\mathcal{L}_{U}$ provide essential complementary gains for modality invariance and code entropy.
% To quantify the specific contribution of each module, we conducted an ablation study on MS COCO as detailed in Table \ref{tab:ablation}. 
% The results establish a distinct performance hierarchy where the metric decreases from 0.7693 with traditional rigid Point Proxies to 0.7615 using the isotropic vMF distribution with $\beta=0$. 
% Crucially, our anisotropic Kent model surpasses both to achieve a peak of 0.7843. This clear performance progression empirically confirms that while probabilistic modeling is beneficial, the capacity to capture \textit{anisotropic} semantic conflicts via shape parameters is the decisive factor in resolving gradient contention. 
% Regarding auxiliary objectives, the Multi-Modal Irrelevance Regularization $\mathcal{L}_{Reg}$ emerges as the second most critical component, evidenced by a performance drop to 0.7326 upon its exclusion. Meanwhile, the Cross-Modal Alignment $\mathcal{L}_{CM}$ and Uniform Constraint $\mathcal{L}_{U}$ provide essential complementary gains for modality invariance and code entropy.

\begin{figure}[t]
\centering
\includegraphics[width=\columnwidth]{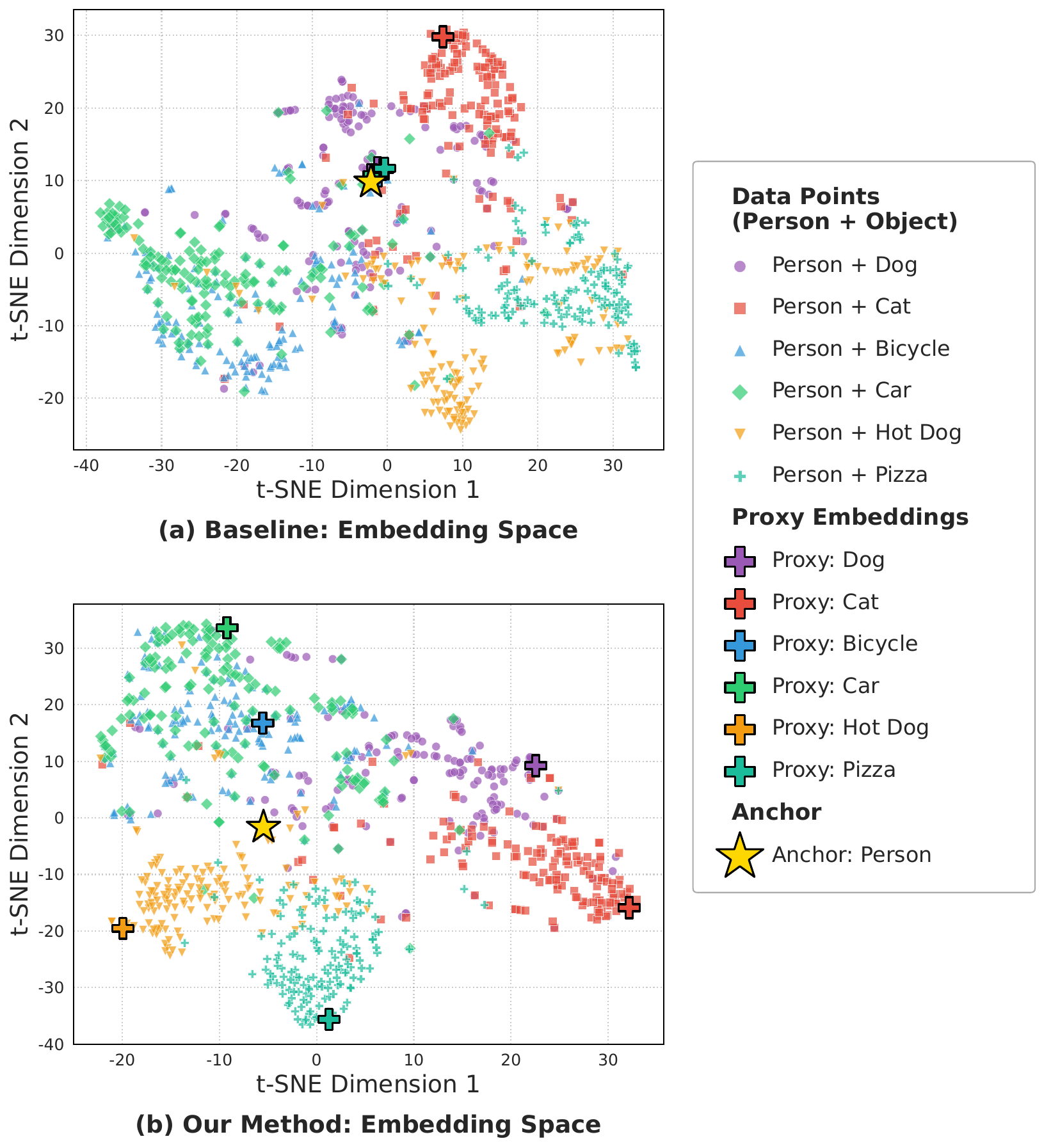}
\caption{t-SNE visualization of the embedding spaces learned by DSPH and KDPH.}
\label{fig:tsne_vis}
\end{figure}
% \vspace{-0.8em}

\subsubsection{Visualizing Distributions Across Multi-label Scenarios}
% We further investigate the robustness of our KDPH approach in topological structures by visualizing the embedding space of the complex Person category on MS COCO using t-SNE. As illustrated in Figure \ref{fig:tsne_vis}a, the Baseline suffers from the challenges related to proxy collapse, where proxies of distinct co-occurring classes physically converge towards the central anchor proxy. In comparison, KDPH (Figure \ref{fig:tsne_vis}b) reveals a distinct semantic-adaptive topology. Benefiting from these powerful components , the embedding space is seamlessly disentangled into coherent radial clusters. The structured architecture demonstrates that our method successfully captures latent semantic hierarchies, dynamically recalibrating decision boundaries to accommodate diverse contextual variations while preserving core identity representations.

To empirically verify the topological superiority of KDPH, we visualized the embedding space of the complex Person category on MS COCO using t-SNE (Figure \ref{fig:tsne_vis}). The Baseline (Fig. \ref{fig:tsne_vis}a) exhibits severe proxy collapse, where proxies of distinct co-occurring classes (e.g., Car, Bicycle) physically collapse onto the central anchor. This visually confirms that rigid point proxies suffer from gradient contention, eliminating decision boundaries. In sharp contrast, KDPH (Fig. \ref{fig:tsne_vis}b) reveals a clear semantic-adaptive topology. The embedding space disentangles into coherent radial clusters based on high-level affinity—grouping animate entities, vehicles, and food along distinct directional axes. This structured layout demonstrates that our anisotropic Kent proxies successfully capture latent semantic hierarchies, accommodating contextual diversity without destabilizing the core identity.

\section{Conclusion}

% In this paper, we address the critical bottleneck of gradient contention in multi-label cross-modal hashing with the Kent-based Distributed Proxy Hashing (KDPH) framework. It uses a decoupled gradient strategy to absorb contextual variance without destabilizing semantic centroids. Experiments show KDPH outperforms state-of-the-art methods, especially in low-bit and high-complexity scenarios, while preserving fine-grained topological structure. Our method achieves these performance gains with no additional inference overhead, offering a robust and efficient solution that paves the way for future advancements in long-tailed recognition and online hashing.
In this paper, we address the critical bottleneck of gradient contention in multi-label cross-modal hashing with the Kent-based Distributional Proxy Hashing (KDPH) framework. By shifting the paradigm from point proxies to anisotropic Kent distributions, our method enables proxies to absorb contextual conflicts by dynamically adjusting directional variance. This geometric flexibility allows the model to maintain stable semantic centroids while stretching to accommodate diverse label correlations. Extensive experiments on three benchmarks demonstrate that KDPH outperforms state-of-the-art methods.

\section*{Acknowledgments}
This work was supported by the Institute of Information Engineer
ing, Chinese Academy of Sciences, under Grant E4101311F3, E4V06811F3. The reimbursement for this work was covered by this grant. 

%% The file named.bst is a bibliography style file for BibTeX 0.99c
\bibliographystyle{named}
\bibliography{ijcai26}

\clearpage      % 关键命令：强制换页并输出所有积压的浮动体(图片/表格)

\pdfpagewidth=8.5in
\pdfpageheight=11in

% ----------------------------------------------------
% 1. 复制原文件的导言区 (Preamble)
% ----------------------------------------------------
% 必须包含 ijcai26 样式包以保持格式一致
% \usepackage{ijcai26}

% Use the postscript times font!
% \usepackage{times}
% \usepackage{soul}
% \usepackage{url}
% \usepackage[utf8]{inputenc}
% \usepackage[small]{caption}
% \usepackage{graphicx}
% \usepackage{amsmath}
% \usepackage{amsthm}
% \usepackage{amssymb}
% \usepackage{booktabs}
% \usepackage{algorithm}
% \usepackage{algorithmic}
% \usepackage[switch]{lineno}
% \usepackage{microtype}
% \usepackage{multirow}
% \usepackage[hidelinks]{hyperref}

% 保持行号设置（如果是提交版本请注释掉）
% \linenumbers

\urlstyle{same}

% 定义定理环境（保持与主文件一致，防止报错）
% \newtheorem{example}{Example}
% \newtheorem{theorem}{Theorem}

% PDF Info (可选，修改为补充材料标题)
\pdfinfo{
/Title (Supplementary Material for KDPH)
/TemplateVersion (IJCAI.2026.0)
}

% ----------------------------------------------------
% 2. 文档主体
% ----------------------------------------------------
% \begin{document}

% 如果需要单独的标题页，可以取消下面注释
% \title{Supplementary Material: Absorbing Gradients Conflicts: Modeling Semantic Variance via Kent Distributions for Cross-Modal Hashing}
% \maketitle

% 声明附录开始
\appendix

% ----------------------------------------------------
% 3. 下面是提取出的附录内容
% ----------------------------------------------------

\section{Extended Analysis for Main Results: Semantic Topological Evolution}
\label{sec:appendix_main_results}

To interpret the quantitative improvements reported in the main results from a topological perspective, we investigate the geometric organization of the learned proxy space. Specifically, we aggregate the 80 fine-grained categories of the MS COCO dataset into 12 semantic super-groups (e.g., \textit{animal, vehicle, electronic, furniture}) based on the standard semantic taxonomy. This allows us to evaluate the \textbf{Semantic Disentanglement} capability of the model by measuring two key metrics: Intra-Group Similarity (the cosine cohesion within a super-group) and Inter-Group Similarity (the cosine overlap between distinct super-groups). An ideal semantic space should maximize the former while minimizing the latter, creating a structured manifold where conceptually related classes cluster tightly and unrelated ones are strictly separated.

\begin{figure*}[t]
    \centering
    % 请确保 png/ 文件夹在当前目录下
    \includegraphics[width=\textwidth]{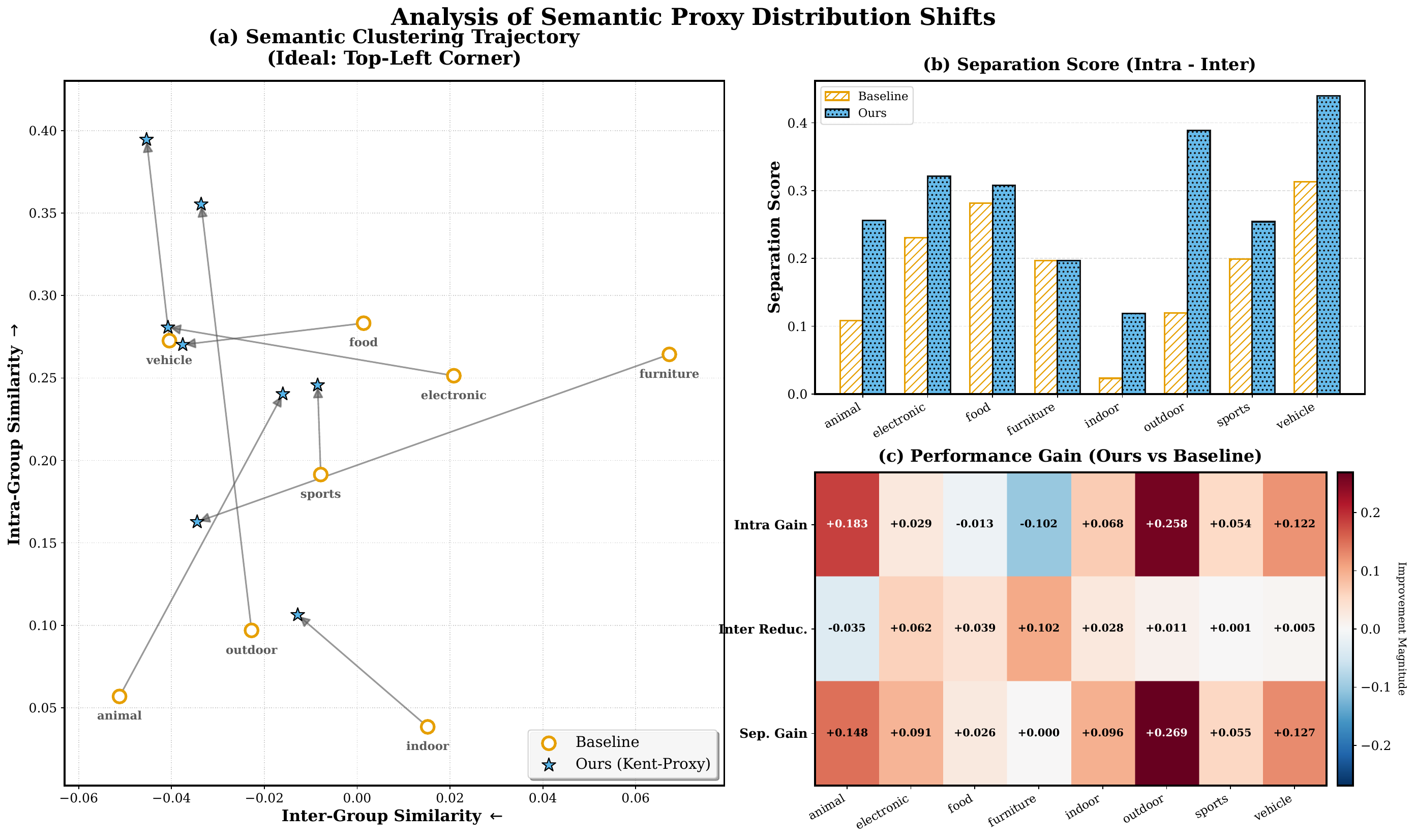}
    \caption{\textbf{Analysis of Semantic Proxy Distribution Shifts.} (a) The scatter plot illustrates the topological quality of semantic super-groups. KDPH (blue stars) consistently migrates towards the top-left Ideal Corner compared to the Baseline (orange circles), indicating higher intra-group cohesion and lower inter-group interference. (b) The Separation Scores confirm that KDPH establishes sharper boundaries between semantic concepts. (c) The heatmap details the specific contributions of Intra-Group Gain and Inter-Group Reduction to the overall performance improvement.}
    \label{fig:semantic_topology}
\end{figure*}

Figure \ref{fig:semantic_topology} visualizes this topological evolution. Subplot (a) plots the semantic trajectory of each super-group in a 2D coordinate system, where the y-axis represents Intra-Group Similarity (higher is better) and the x-axis represents Inter-Group Similarity (lower is better). The Baseline proxies (orange circles) are scattered loosely across the space, indicating a high degree of semantic entanglement where distinct concepts (e.g., \textit{furniture} and \textit{electronic}) share excessive similarity. In contrast, the KDPH proxies (blue stars) exhibit a distinct Top-Left Migration. The gray arrows highlight this consistent shift towards the Ideal Corner (high cohesion, low interference). For instance, the \textit{vehicle} and \textit{outdoor} groups show dramatic improvements in separation, verifying that our anisotropic distribution effectively pushes irrelevant semantic clusters apart.

Subplots (b) and (c) further quantify these gains. The Separation Score, defined as the difference between Intra- and Inter-group similarities, is consistently higher for KDPH across nearly all categories. The heatmap in Subplot (c) decomposes this improvement, revealing that KDPH achieves this separation primarily through a substantial reduction in Inter-Group similarity (blue cells in the middle row) and a simultaneous boost in Intra-Group cohesion (red cells in the top row). This confirms that the performance gains observed in Table \ref{tab:comparison_sota_full} stem from a fundamentally better-structured semantic embedding space, where the distributed proxies enforce a rigorous hierarchy that aligns with human conceptual understanding.

\begin{figure}[h]
\centering
\includegraphics[width=\columnwidth]{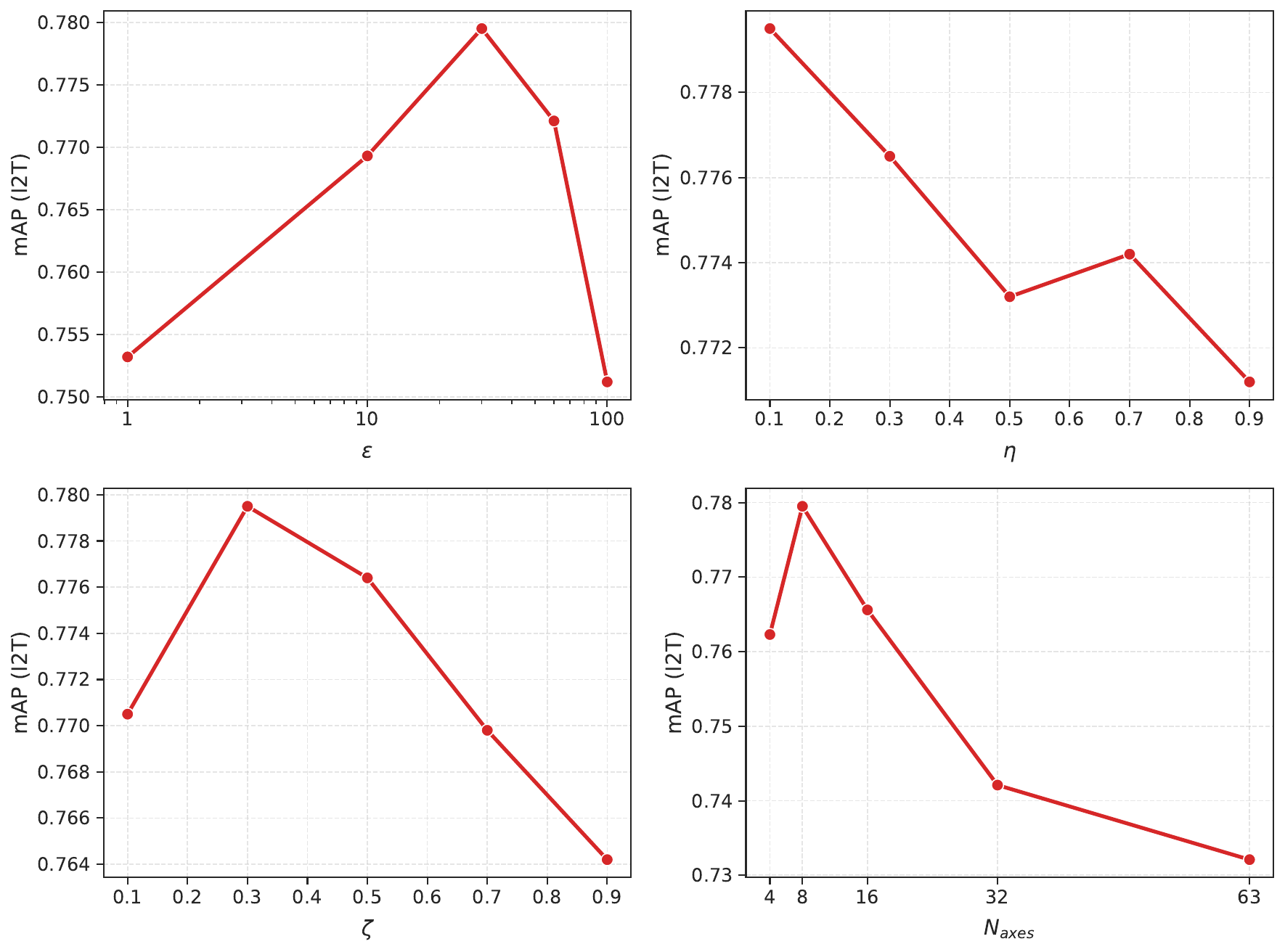}
\caption{\textbf{Parameter Sensitivity Analysis on MS COCO (@64 bits).} We evaluate the impact of the loss weights $\epsilon, \eta, \zeta$ and the number of principal axes $N_{axes}$ on the Image-to-Text (I2T) mAP. The results justify our default settings ($\epsilon=40, \eta=0.1, \zeta=0.3, N_{axes}=8$).}
\label{fig:sensitivity}
\end{figure}

\section{Extended Analysis for Proxy Oscillation and Stability}
\label{sec:appendix_rq1}

To substantiate the robustness of the findings presented in RQ1 (Can our proposed model effectively mitigate the 'Chaos Proxy Oscillation' problem?), we provide a comprehensive visualization of training dynamics across a broader spectrum of categories. While the main text highlights representative cases, this appendix expands the analysis to twelve additional classes from the MS COCO dataset. These categories are stratified into high and low co-occurrence groups to demonstrate the universal effectiveness of the KDPH framework across varying degrees of semantic complexity.

\subsection{Mitigation of Oscillation in Complex Scenarios}
\begin{figure*}[t]
    \centering
    \includegraphics[width=\textwidth]{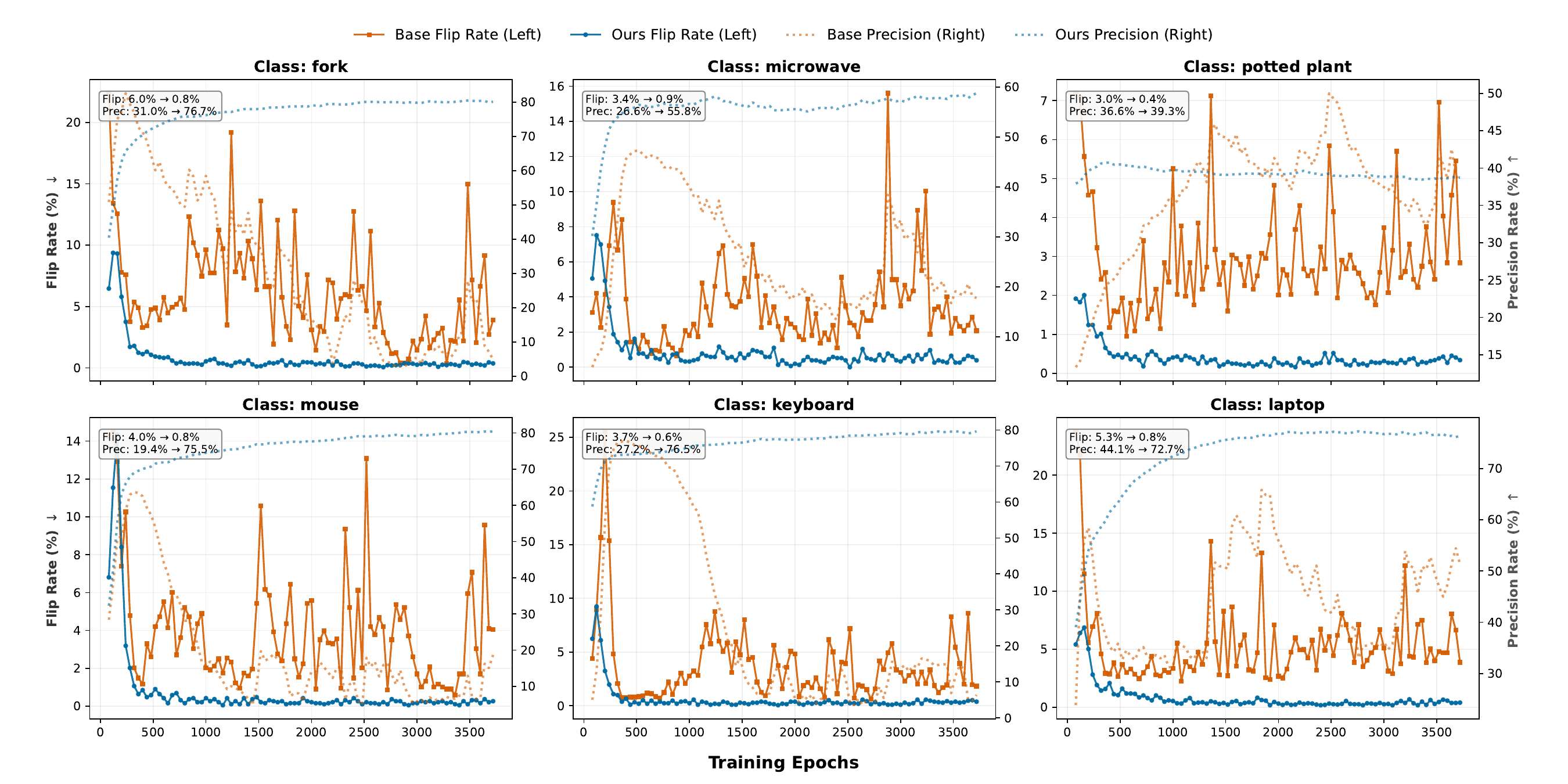}
    \caption{\textbf{Supplementary for RQ1: Training Dynamics on High Co-occurrence Classes.} The Baseline (orange) exhibits chaotic oscillation (high flip rates) and jagged precision curves due to gradient contention. KDPH (blue) achieves rapid stabilization and superior convergence, validating its ability to resolve semantic conflicts.}
    \label{fig:supp_high}
\end{figure*}

Figure \ref{fig:supp_high} delineates the training trajectories for complex, high co-occurrence categories such as \textit{fork, microwave, potted plant, mouse, keyboard,} and \textit{laptop}. In these scenarios, dense semantic overlaps typically induce severe gradient contention for traditional methods. Observing the baseline metrics (orange curves), it is evident that rigid point proxies suffer from persistent oscillation, with the Proxy Flip Rate fluctuating dangerously between $5\%$ and $20\%$ even in late training stages. This instability directly correlates with the jagged and suboptimal average precision curves, confirming that static point representations fail to anchor semantically ambiguous samples. In sharp contrast, the proposed KDPH framework (blue curves) effectively decouples these conflicting gradients, suppressing the flip rate to near-zero levels almost immediately. The resulting precision curves exhibit a smooth, monotonic ascent, confirming that the anisotropic Kent distributions successfully absorb the variance that otherwise destabilizes point-based models.

\subsection{Preservation of Stability in Simple Scenarios}
\begin{figure*}[t]
    \centering
    \includegraphics[width=\textwidth]{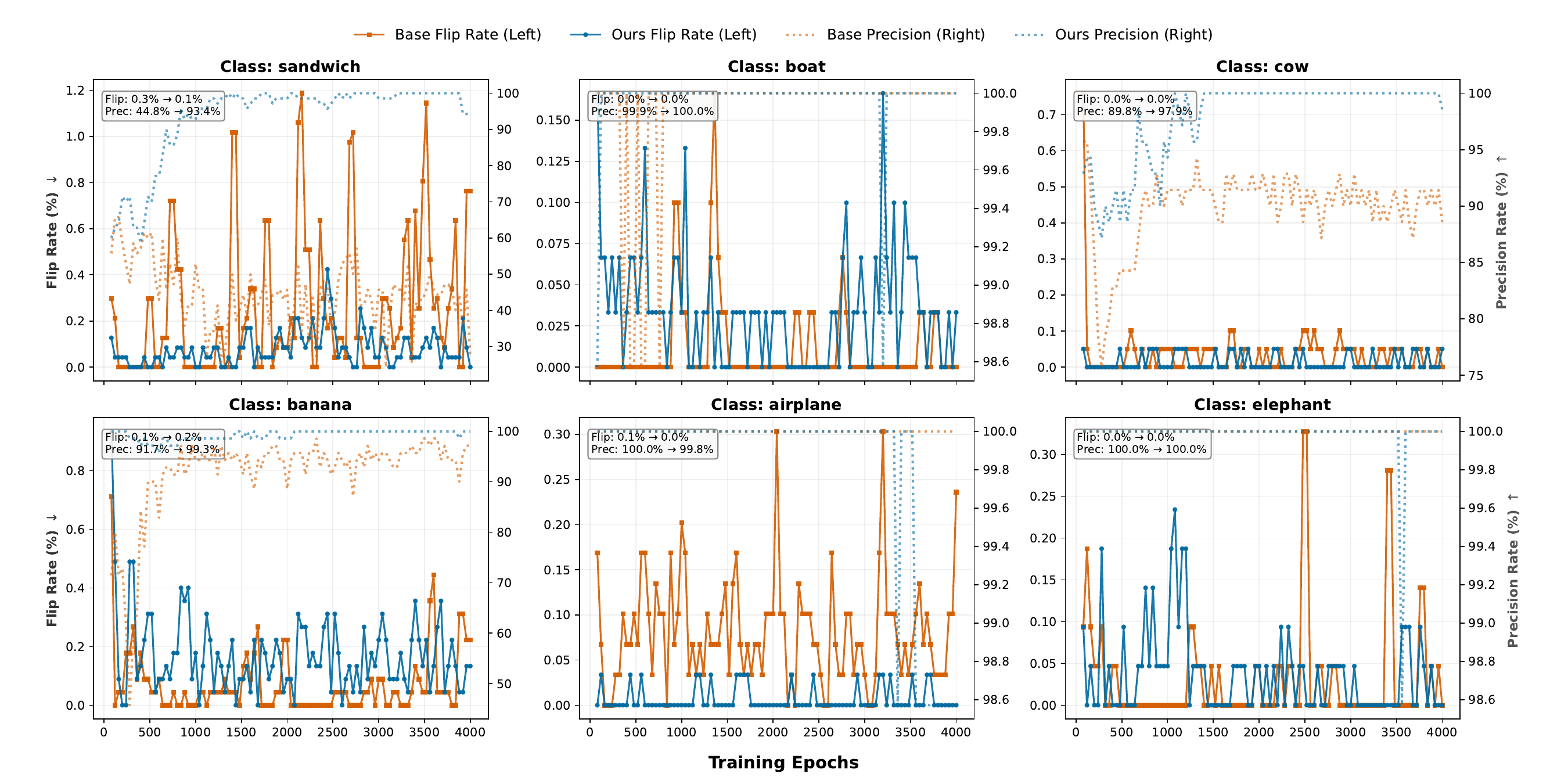}
    \caption{\textbf{Supplementary for RQ1: Training Dynamics on Low Co-occurrence Classes.} Both methods maintain high stability. This confirms that KDPH's anisotropic modeling is robust and does not negatively impact learning in simple, distinct semantic scenarios.}
    \label{fig:supp_low}
\end{figure*}

Figure \ref{fig:supp_low} illustrates the dynamics for semantically distinct, low co-occurrence categories, including \textit{sandwich, boat, cow, banana, airplane,} and \textit{elephant}. In these semantically pure environments, the baseline method performs adequately with minimal oscillation. Crucially, KDPH mirrors this stability, maintaining low flip rates and high precision without introducing unnecessary variance or computational noise. This comparative analysis serves as a vital validation of robustness: it demonstrates that the distributed proxy mechanism is adaptive, providing necessary topological flexibility for complex classes while behaving like a stable, confident estimator in unambiguous scenarios.

% -------------------------------------------------------------------
% Figures for Appendix
% -------------------------------------------------------------------

\subsection{Quantitative Correlation between Proxy Chaos and Performance}
To move beyond visual inspection and quantitatively verify the causal link between proxy oscillation and retrieval degradation, we introduce a composite metric named the \textbf{Chaos Index}. This scalar, normalized between 0 and 1, aggregates four kinematic properties of the proxy's learning trajectory: \textit{Path Efficiency} (ratio of displacement to total path length), \textit{Directional Consistency} (cosine similarity between consecutive steps), \textit{Movement Entropy} (randomness of step magnitudes), and \textit{Autocorrelation} (predictability based on vector dot products). The Chaos Index is computed by averaging the normalized entropy with the inverted values of the three ordered metrics, where a value of 1 indicates maximum stochasticity. We also define the Average Co-occurring Labels ($ACL$) to quantify semantic complexity: $ACL(c) = \frac{1}{|N_c|} \sum_{i \in N_c} (|l_i|_1 - 1)$, representing the mean number of concurrent labels associated with class $c$.

\begin{figure*}[t]
    \centering
    \includegraphics[width=\textwidth]{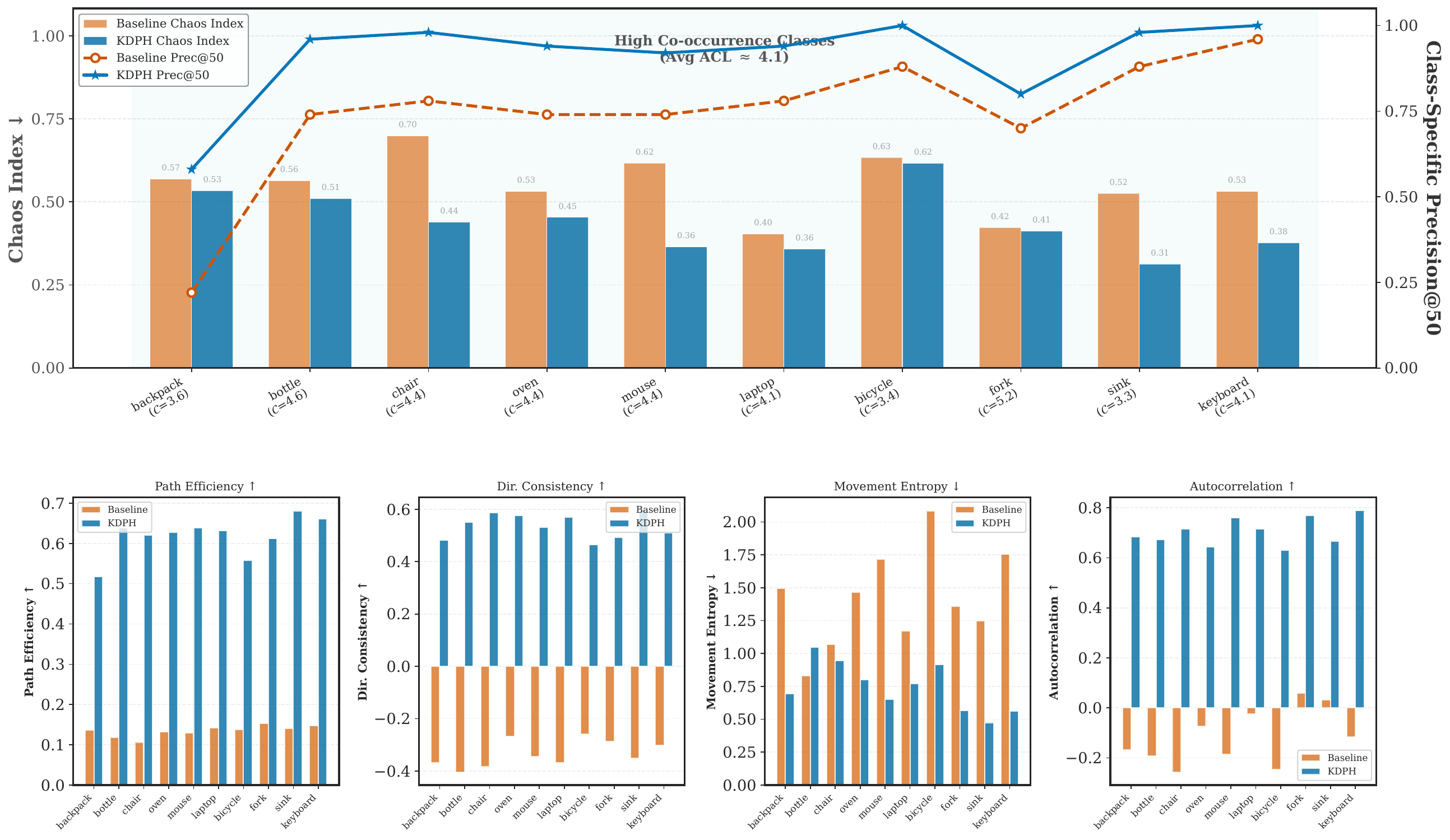}
    \caption{\textbf{Chaos Index vs. Real Performance (High Co-occurrence Classes).} The charts correlate kinematic stability (Chaos Index, bars) with retrieval accuracy (Precision@50, lines). For complex classes (Avg ACL $\approx$ 4.1), the Baseline suffers from high chaos and low precision. KDPH effectively suppresses this chaos, resulting in higher path efficiency and significantly improved retrieval performance.}
    \label{fig:chaos_high}
\end{figure*}

Figure \ref{fig:chaos_high} presents the analysis for high-complexity classes (Avg $ACL \approx 4.1$) on the MS COCO dataset (64 bits). A distinct inverse correlation is observed between the Chaos Index and retrieval performance (Precision@50). The baseline method (orange bars) exhibits elevated Chaos Indices across all dense categories, driven by low path efficiency and high movement entropy. This kinematic instability directly suppresses the precision (orange dotted line). Conversely, KDPH (blue bars) significantly minimizes the Chaos Index, transforming the erratic trajectory into a directed, efficient optimization path. This stabilization translates directly into substantial performance gains, as evidenced by the superior Precision@50 scores (blue star line), confirming that resolving gradient contention is the underlying mechanism behind our performance improvements.

\begin{figure*}[t]
    \centering
    \includegraphics[width=\textwidth]{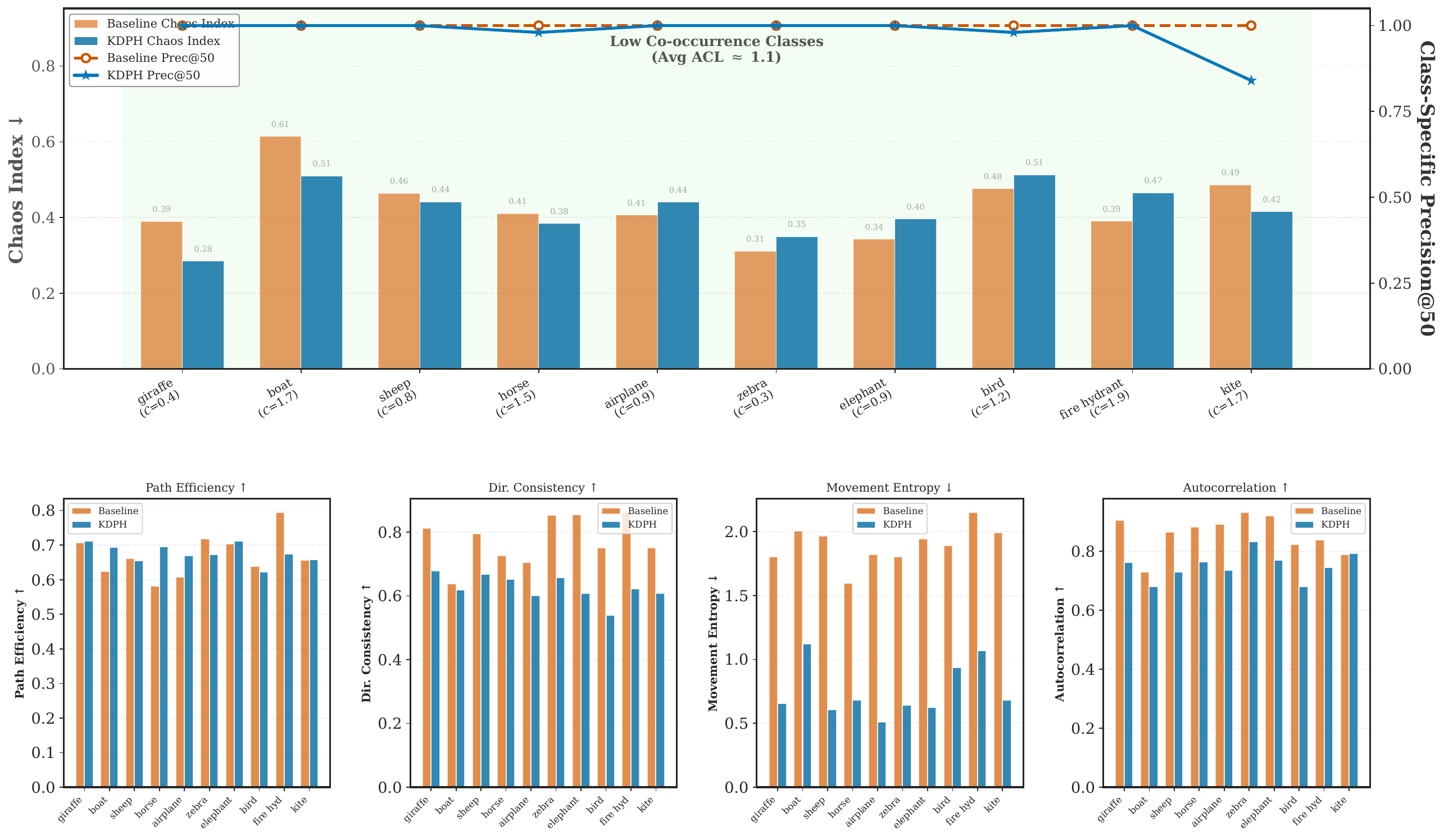}
    \caption{\textbf{Chaos Index vs. Real Performance (Low Co-occurrence Classes).} In scenarios with low semantic interference (Avg ACL $\approx$ 1.1), both methods maintain low Chaos Indices and high Autocorrelation. This demonstrates that KDPH retains the optimization efficiency of point-based methods when gradient contention is absent.}
    \label{fig:chaos_low}
\end{figure*}

In parallel, Figure \ref{fig:chaos_low} details the metrics for low-complexity classes (Avg $ACL \approx 1.1$). In these semantically sparse scenarios, the gradient signals are naturally coherent. Consequently, both the Baseline and KDPH exhibit low Chaos Indices and high, comparable Autocorrelation scores. The overlapping performance curves further reinforce our conclusion: the proposed anisotropic Kent distribution effectively reduces to a stable estimator in simple scenarios, only activating its shape-deforming capacity when high variance demands it. This confirms that KDPH provides an improvement without inference overhead, offering robustness in complex regimes without compromising the intrinsic stability of simple categories.

\section{Extended Analysis for Quantifying Proxy Collapse via GTSD}
\label{sec:appendix_rq2}

To provide a fine-grained perspective on the Proxy Collapse phenomenon discussed in RQ2, we introduce the \textbf{Ground-Truth Score Divergence (GTSD)}. While retrieval metrics like mAP measure global ranking, GTSD specifically quantifies the severity of the gradient competition within a single multi-label instance. It measures the disparity between the model's confidence in the best-recognized label versus the worst-recognized label. Formally, for a sample $x$ with a ground-truth label set $\mathcal{L}_x = \{l_1, ..., l_K\}$ and a scoring function $S(x, c)$, GTSD is defined as:
\begin{equation}
    \text{GTSD}(x) = \left( \max_{l_i \in \mathcal{L}_x} S(x, l_i) \right) - \left( \min_{l_j \in \mathcal{L}_x} S(x, l_j) \right)
\end{equation}
A high GTSD indicates that the model is collapsing onto a subset of dominant labels (head classes) while suppressing subordinate ones (tail classes). Conversely, a low GTSD implies intra-sample equity, where the proxy mechanism successfully maintains high compatibility for all co-occurring semantic concepts simultaneously.

\begin{figure*}[t]
    \centering
    % 请确保文件名路径正确
    \includegraphics[width=\textwidth]{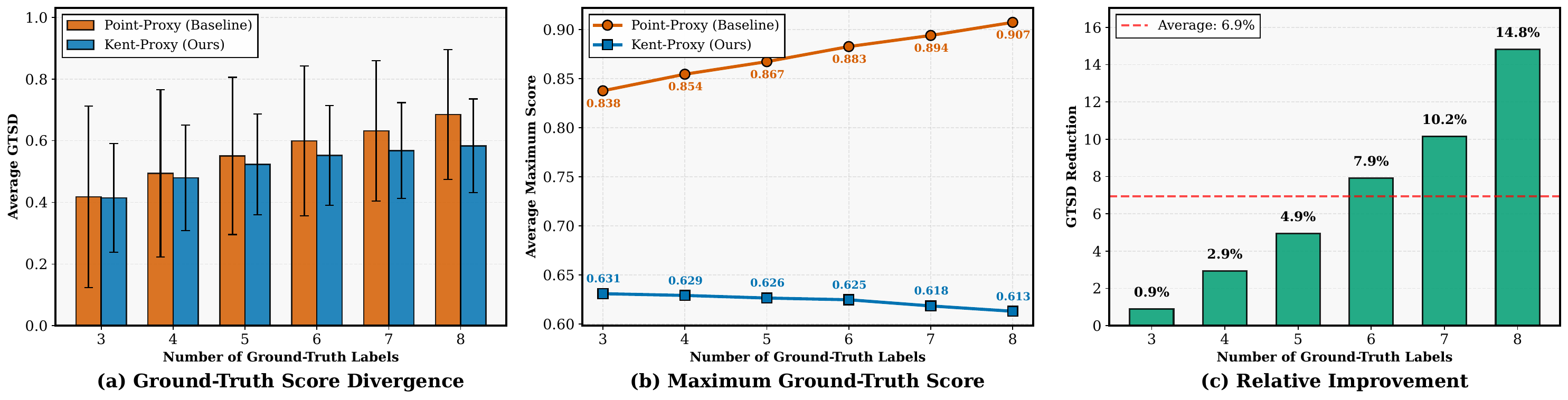}
    \caption{\textbf{Ground-Truth Score Divergence (GTSD) Analysis.} (a) As the number of labels increases, the Baseline exhibits a widening gap between the best and worst scored labels, indicating proxy collapse. KDPH maintains a stable, lower divergence. (b) The Baseline's high maximum scores suggest overfitting to dominant labels. (c) KDPH demonstrates increasing efficacy in complex scenarios, reducing the divergence by up to $14.8\%$ in dense multi-label settings.}
    \label{fig:gtsd_analysis}
\end{figure*}

Figure \ref{fig:gtsd_analysis} elucidates the internal scoring dynamics on the MS COCO dataset as the semantic complexity (number of labels) increases from 3 to 8. Subplot (a) reveals that the Baseline (orange) suffers from an exacerbated divergence as the label density grows. This confirms that rigid point proxies, unable to satisfy conflicting gradients, prioritize the easiest category at the expense of others. In contrast, KDPH (blue) maintains a significantly tighter score distribution. Even more telling is Subplot (b), which shows that the Baseline achieves higher \textit{maximum} scores ($>0.9$) compared to KDPH ($\approx 0.6$). This is not indicative of superior performance, but rather of \textbf{overconfidence and collapse}: the point proxy is pulled entirely towards the dominant centroid, abandoning the geometric structure required for the remaining labels.

Subplot (c) quantifies the relative improvement of KDPH over the Baseline in reducing GTSD. The benefit of our anisotropic modeling scales broadly with complexity, achieving a $14.8\%$ reduction in divergence for samples with 8 labels. This empirically proves that the Kent-based distribution effectively utilizes its shape parameters to cover the semantic manifold of multiple labels, preventing the topological collapse that characterizes traditional point-based hashing.

\section{Hyperparameter Sensitivity Analysis}
\label{sec:appendix_sensitivity}

To verify the robustness of KDPH and investigate the impact of different loss components and geometric constraints, we conducted a detailed sensitivity analysis on the MS COCO dataset with a hash code length of 64 bits. We specifically examined four critical hyperparameters: the weights for the regularization term ($\epsilon$), the cross-modal alignment term ($\zeta$), and the uniform distribution constraint ($\eta$), as well as the number of principal axes ($N_{axes}$) in the Kent distribution. The results are illustrated in Figure \ref{fig:sensitivity}.

\subsection{Impact of Loss Coefficients}
\paragraph{Irrelevance Regularization ($\epsilon$).} 
As shown in the top-left plot of Figure \ref{fig:sensitivity}, the model's performance exhibits a significant sensitivity to the Multi-Modal Irrelevance Regularization weight $\epsilon$. The mAP improves steadily as $\epsilon$ increases from 1 to 40, confirming that strictly penalizing high similarities between semantically disjoint pairs is crucial for sharpening decision boundaries. However, setting $\epsilon$ too high (e.g., 100) causes a sharp performance degradation. We hypothesize that excessive regularization dominates the gradient optimization, suppressing the learning of positive semantic correlations. Thus, $\epsilon=40$ strikes the optimal balance.

\paragraph{Uniform Constraint ($\eta$).} 
The top-right plot illustrates the effect of the Optimal Transport-based uniform constraint. We observe that a relatively small weight ($\eta=0.1$) yields the best performance. As $\eta$ increases, the performance generally declines. This suggests that while maximizing code entropy is beneficial for preventing bit collapse, over-enforcing uniformity may distort the intrinsic semantic manifold, forcing the embeddings into an unnatural distribution that hinders discrimination. Therefore, we treat this as a soft constraint with $\eta=0.1$.

\paragraph{Cross-Modal Alignment ($\zeta$).} 
The bottom-left plot analyzes the contribution of the contrastive alignment loss. The performance peaks at $\zeta=0.3$. While cross-modal alignment is essential for bridging the modality gap, an overly large $\zeta$ might cause the model to focus excessively on instance-level matching at the expense of class-level semantic structure learned by the proxy module.

\subsection{Impact of Anisotropic Geometry ($N_{axes}$)}
A unique contribution of KDPH is the modeling of anisotropic variance via the Kent distribution's principal axes. The bottom-right plot in Figure \ref{fig:sensitivity} investigates the optimal number of axes $N_{axes}$ to model the shape parameter $\boldsymbol{\beta}$.
Interestingly, the performance does not increase monotonically with the number of axes. The best result is achieved at $N_{axes}=8$ (which corresponds to $K/8$ for 64-bit codes). 
\begin{itemize}
    \item When $N_{axes}$ is too small (e.g., 4), the proxy lacks sufficient degrees of freedom to capture complex, multi-directional semantic variances.
    \item Conversely, increasing $N_{axes}$ beyond 16 (e.g., 32 or 63) leads to a substantial drop in accuracy. This phenomenon can be attributed to the curse of dimensionality and overfitting: utilizing too many shaping parameters allows the proxy to model noise rather than meaningful semantic directions, destabilizing the optimization.
\end{itemize}
This finding validates our choice of $N_{axes}=K/8$ as a parsimonious yet effective configuration for modeling anisotropic semantic manifolds in the Hamming space.

% \end{document}

\end{document}